\documentclass[10pt,twocolumn,letterpaper]{article}

\usepackage[pagenumbers]{cvpr}      

\usepackage{multirow}

\definecolor{cvprblue}{rgb}{0.21,0.49,0.74}
\usepackage[pagebackref,breaklinks,colorlinks,allcolors=cvprblue]{hyperref}

\def\confName{CVPR}
\def\confYear{2027}

\title{Speculate in the Right Order: Efficient Self-Speculative Decoding for Autoregressive Image Generation}

\author{
  Yihan Chen \quad
  Jintao Zhang \quad
  Jun Zhu\thanks{Corresponding author:
  \href{mailto:dcszj@mail.tsinghua.edu.cn}{dcszj@mail.tsinghua.edu.cn}.}\\
  Tsinghua University 
}

\begin{document}
  \maketitle

  \begin{abstract}
Self-speculative decoding accelerates autoregressive
image generation by leveraging the same model for
parallel drafting and sequential verification,
without requiring an additional draft model.
While existing methods primarily focus on improving
drafting and verification strategies, recent
advances in random-order
autoregressive modeling introduce an additional degree of freedom:
\textbf{the generation order.}
In this work, we theoretically analyze how generation order affects self-speculative decoding efficiency, identifying two key factors:
draft--target mismatch, which determines token
rejection probabilities, and early-rejection cost,
which reflects the higher cost of rejecting tokens earlier in a speculative round.
Motivated by these insights, we propose \textbf{Speculation-Aware Ordering (SAO)}, a training-free framework for offline
optimization of generation orders.
Instead of evaluating candidate orders through costly speculative decoding, SAO uses simple pairwise conditional-TV statistics to formulate a rejection-cost-aware objective and efficiently construct speculation-friendly generation orders offline.
Extensive experiments across different backbones and speculative widths demonstrate that SAO improves decoding efficiency, with the additional benefit of improved generation quality.
Furthermore, SAO provides complementary efficiency gains when combined with existing self-speculative decoding methods.
To the best of our knowledge, this is the first work to identify generation order as a new optimization dimension for efficient self-speculative decoding in autoregressive image generation.
\end{abstract}
  \section{Introduction}
\label{sec:intro}
Autoregressive (AR) models have achieved remarkable success in text generation~\cite{radford2019language,brown2020language} and have been widely extended to image generation~\cite{esser2021taming,sun2024autoregressive}. 
Traditional AR image generation methods  typically generate visual tokens sequentially in a fixed order, leading to high inference latency.
Recent advances in random-order AR models, such as RandAR~\cite{pang2025randar} and ARPG~\cite{li2026arpg}, introduce flexibility in generation order and support parallel token prediction, opening new opportunities for
efficient image generation.

Although parallel decoding reduces inference latency, predicting multiple visual tokens simultaneously neglects their intra-group dependencies, potentially degrading generation quality~\cite{zhang2026locality}. 
To address this limitation, speculative decoding combines parallel token drafting with sequential verification and rejection correction, enabling acceleration while preserving the target autoregressive distribution~\cite{leviathan2023speculative,chen2023speculative}.
In particular, self-speculative decoding leverages the same AR model for both drafting and verification, eliminating the need for an additional draft model~\cite{teng2025sjd,guo2026assd}.
Existing self-speculative decoding methods primarily improve speculative decoding through better draft construction, sampling, and verification strategies~\cite{teng2025sjd,so2026scd,so2025gsd}. 
Beyond these directions, random-order AR models offer an additional degree of freedom: \textbf{the generation order. }
This raises a natural question: Can we optimize the generation order to further improve self-speculative decoding efficiency?

\begin{figure}[t]
      \centering
      \includegraphics[width=\columnwidth]{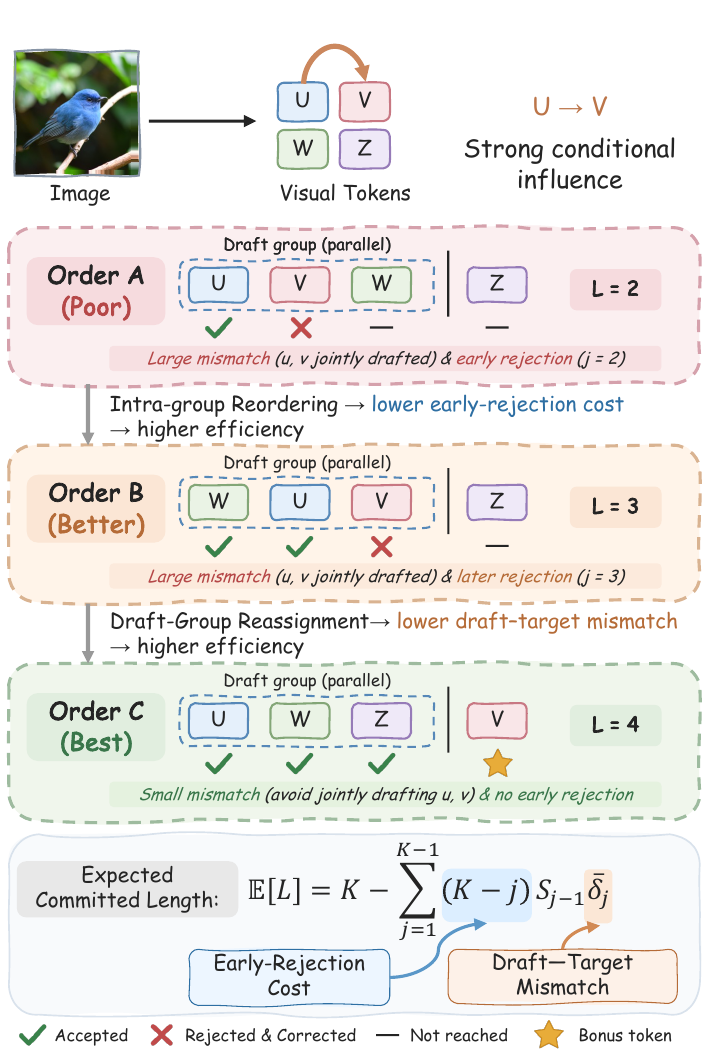}
      \caption{
      A toy example illustrating how generation order affects
      self-speculative decoding.
      We consider four visual tokens, where $u$ has a strong conditional
      influence on $v$, while other pairwise influences are assumed
      negligible.
      With $K=4$, the illustrated round uses three parallel draft proposals
      and a potential bonus token.
      From Order A to B, intra-group reordering delays the illustrated
      rejection from $j=2$ to $j=3$, reducing the early-rejection cost.
      From Order B to C, placing $v$ at the bonus position avoids jointly
      drafting $u$ and $v$, potentially reducing the draft--target mismatch
      associated with their dependency.
      The committed lengths ($L=2,3,4$) represent illustrative decoding
      outcomes.
      }
      \label{fig:motivation}
\end{figure}

As illustrated by the toy example in
\cref{fig:motivation}, generation order affects self-speculative decoding through draft--target mismatch and early-rejection cost:
\textbf{(1) Draft--Target Mismatch.} Different generation orders induce different mismatches between draft and target distributions, affecting token acceptance probabilities~\cite{leviathan2023speculative,so2025gsd}.
\textbf{(2) Early-Rejection Cost.} The position of the first rejection determines how many tokens can be committed, with earlier rejections resulting in lower efficiency.
Therefore, an effective generation order should minimize draft--target mismatch and avoid costly early rejections.
However, identifying such an order is challenging,
as directly evaluating the decoding efficiency
of numerous candidate orders is computationally expensive.

To address this challenge, we propose
\textbf{Speculation-Aware Ordering (SAO)}, a training-free framework for optimizing generation orders offline to improve self-speculative decoding efficiency.
We first estimate pairwise conditional
influences between visual tokens using
conditional total variation (TV).
Based on these simple statistics, we formulate a rejection-cost-aware objective that incorporates both draft--target mismatch and early-rejection cost.
Finally, exploiting the pairwise additive structure of this objective, we employ the method of conditional expectations to efficiently construct a complete generation order.

Our main contributions are summarized as follows:
\begin{itemize}
    \item To the best of our knowledge, we are the first to identify generation order as a new optimization dimension for efficient self-speculative decoding in autoregressive image generation, and theoretically analyze its impact through draft--target mismatch and early-rejection cost.

    \item We propose Speculation-Aware Ordering (SAO), a training-free framework that optimizes generation orders offline using simple conditional statistics and a rejection-cost-aware ordering objective.

    \item We demonstrate the effectiveness of SAO across different speculative widths and model backbones, achieving consistent efficiency improvements, additional gains in generation quality, and complementarity with existing self-speculative decoding methods.
\end{itemize}

  \section{Theoretical Analysis}
\label{sec:theoretical_analysis}

\subsection{Self-Speculative Decoding with Flexible Generation Orders}
\label{sec:self_spec}

\noindent\textbf{Arbitrary-Order AR Modeling.}
Given an image represented by $N$ visual tokens $\mathbf{x} = (x_1, \ldots, x_N)$, a random-order AR model supports generation under an arbitrary
permutation $\pi \in \mathcal{S}_N$~\cite{pang2025randar,li2026arpg}.
The corresponding joint distribution is factorized as
\begin{equation}
p_\theta^\pi(\mathbf{x}\mid\ell)
=
\prod_{i=1}^{N}
p_\theta(x_{\pi_i}\mid c_{i-1}^{\pi}),
\end{equation}
where $c_{i-1}^{\pi}$ denotes the ordered generation
history under permutation $\pi$, including the conditioning information $\ell$ (e.g., a class label or a text prompt).
Unlike conventional AR models with a fixed token order, random-order AR models allow $\pi$ to be selected
without retraining the model.

\noindent\textbf{Self-Speculative Decoding.}
Given an ordered generation history $c_t=(\ell,(\pi_1,x_{\pi_1}),\ldots,
(\pi_t,x_{\pi_t}))$
and a speculative window of size $K$, we consider the next $K$ positions in the target order.
Following standard speculative decoding~\cite{leviathan2023speculative,chen2023speculative}, each round consists of $K-1$ draft tokens and one potential bonus token.
Using the same AR model, we independently sample $K-1$ draft tokens in parallel:
\begin{equation}
y_j \sim q_j,\quad
q_j(a)=p_\theta(x_{\pi_{t+j}}=a\mid c_t),
\quad j=1,\ldots,K-1.
\end{equation}

The draft tokens are then verified against the
sequentially conditioned target distributions:
\begin{equation}
p_j(a)=p_\theta(x_{\pi_{t+j}}=a\mid c_t,y_{<j}).
\end{equation}
Conditioned on all preceding drafts being accepted,
the $j$-th draft token is accepted with probability
\begin{equation}
\alpha_j(y_j)
=
\min\left(1,\frac{p_j(y_j)}{q_j(y_j)}\right).
\end{equation}
Upon the first rejection, a corrected token is sampled from
\begin{equation}
p_j^{\mathrm{corr}}(a)
\propto
\max(p_j(a)-q_j(a),0).
\end{equation}
The decoder commits the accepted draft prefix and the
corrected token.
If all drafts are accepted, an additional bonus token
is sampled from the target distribution conditioned on
all accepted drafts, yielding $K$ committed tokens.
Following the exact sampling guarantee established by ASSD~\cite{guo2026assd}, this procedure preserves the target distribution $p_\theta^\pi$ for any fixed generation order $\pi$.


\subsection{Draft--Target Mismatch}
\label{sec:mismatch}
In self-speculative decoding, parallel drafting predicts multiple tokens conditioned on the same prefix, while sequential verification additionally
conditions on preceding accepted drafts.
Generation order \(\pi\) determines which tokens are drafted together and which preceding tokens are omitted from the draft distribution, thereby affecting draft--target mismatch.

We characterize this mismatch using total variation (TV) distance.
Under independent parallel drafting, conditioned on the generated prefix and previously accepted drafts, the expected acceptance probability of the $j$-th token is~\cite{leviathan2023speculative,so2025gsd}
\begin{equation}
\begin{aligned}
\mathbb{E}_{y_j \sim q_j}[\alpha_j(y_j)]
&= \sum_a \min(p_j(a), q_j(a)) \\
&= 1 - \operatorname{TV}(p_j, q_j),
\end{aligned}
\end{equation}
where
$\operatorname{TV}(p_j,q_j)
= \frac{1}{2}\sum_a |p_j(a)-q_j(a)|$.

Consequently, the conditional rejection probability is
\begin{equation}
\begin{aligned}
\delta_j(c_t,y_{<j})
&\triangleq \operatorname{TV}(p_j,q_j)\\
&=\Pr(\mathrm{Reject}_j
\mid c_t,y_{<j},A_{<j}),
\end{aligned}
\label{eq:rejection_tv}
\end{equation}
where $A_{<j}$ denotes the event that all preceding
drafts are accepted.

Therefore, generation order directly affects token
rejection probabilities through draft--target mismatch.

\subsection{Early-Rejection Cost}
\label{sec:rejection_cost}
In self-speculative decoding, only consecutive accepted drafts can be committed, and the first rejection causes
all subsequent draft proposals to be discarded.
Therefore, the position of the first rejection directly
affects the number of committed tokens.

Unless otherwise specified, all probabilities and
expectations are conditioned on the current generation
history $c_t$ and target order $\pi$.
Let $R \in \{1,\ldots,K\}$ denote the first-rejection position within a speculative round, where $\Pr(R=j)$ is the probability that the $j$-th draft is the first rejected token, and
$R=K$ indicates that all $K-1$ drafts are accepted.
Upon rejection at position $j<K$, a corrected token
is committed, yielding $L=j$; otherwise, $L=K$.
Thus, the expected committed length is
\begin{equation}
\begin{aligned}
\mathbb{E}[L]
&= \sum_{j=1}^{K-1} j\,\Pr(R=j)
   + K\Pr(R=K) \\
&= K-\sum_{j=1}^{K-1}(K-j)\Pr(R=j).
\end{aligned}
\end{equation}

To relate the first-rejection probability to
draft--target mismatch, we define the prefix survival
probability $S_{j-1}=\Pr(A_{<j})$, with $S_0=1$,
and the expected conditional mismatch
\begin{equation}
\bar{\delta}_j
=
\mathbb{E}\left[
\delta_j(c_t,y_{<j})\mid A_{<j}
\right].
\end{equation}

Then, the first-rejection probability satisfies
\begin{equation}
\Pr(R=j)=S_{j-1}\bar{\delta}_j.
\end{equation}

Substituting this into the expected committed length,
we obtain
\begin{equation}
\mathbb{E}[L]
=
K-\sum_{j=1}^{K-1}
\underbrace{(K-j)}_{\substack{\text{Early-Rejection}\\\text{Cost}}}
S_{j-1}
\underbrace{\bar{\delta}_j}_{\substack{\text{Draft--Target}\\\text{Mismatch}}}.
\label{eq:expected_length_tv}
\end{equation}

This formulation reveals two factors governing speculative efficiency: draft--target mismatch determines rejection probabilities, while early-rejection cost determines the impact of rejection positions on generation progress.

\subsection{Problem Formulation}
\label{sec:problem_formulation}

Let $T_\pi$ denote the number of speculative rounds
required to generate $N$ visual tokens under target
order $\pi$, and $L_r^\pi$ denote the committed length
of round $r$.
Since
\begin{equation}
\sum_{r=1}^{T_\pi}L_r^\pi=N,
\end{equation}
longer committed prefixes generally reduce the
number of speculative rounds.

Our objective is therefore to find a
generation order that minimizes the expected
number of speculative rounds:
\begin{equation}
\pi^*
=
\arg\min_{\pi\in\mathcal{S}_N}
\mathbb{E}[T_\pi].
\label{eq:ordering_problem}
\end{equation}

Since directly optimizing this objective is computationally expensive, we construct a tractable surrogate objective that accounts for both draft--target mismatch and early-rejection cost.

\section{Method}
\label{sec:method}

We propose Speculation-Aware Ordering (SAO), a training-free framework that constructs speculation-friendly generation orders entirely offline.

\subsection{Conditional-TV Proxy}
\label{sec:conditional_tv}

To characterize pairwise conditional influence between
visual tokens, we introduce a conditional-TV proxy
that measures how revealing one token changes the
prediction of another.
For two distinct positions $u$ and $v$, we define
\begin{equation}
\begin{aligned}
W_{u \rightarrow v}
=
\mathbb{E}_{h \sim \mathcal{D}^{uv}}
\mathbb{E}_{a \sim p_\theta(x_u \mid h)}
\big[
\operatorname{TV}(
&p_\theta(x_v \mid h),\\
&p_\theta(x_v \mid h \oplus (u,a)))
\big],
\end{aligned}
\end{equation}
where $h$ denotes an ordered generation history
containing the conditioning information and previously
generated token positions and values.
The operation $h \oplus (u,a)$ appends token value $a$
at position $u$ to the history.
Here, $\mathcal{D}^{uv}$ denotes the distribution
of sampled generation histories in which neither
$u$ nor $v$ has been generated.
A larger $W_{u \rightarrow v}$ indicates a greater
expected change in the prediction of token $v$
upon revealing token $u$.

To estimate $W$, we sample generation histories of varying prefix lengths from the frozen AR generator under random generation orders.
For each eligible position pair $(u,v)$,
we sample a token value $a$ at position $u$ and measure the resulting change in the predictive distribution of token $v$ using TV.
Averaging these measurements over eligible histories for each pair yields a single global influence matrix $W$, which is subsequently used for offline generation-order optimization.

\subsection{Rejection-Cost-Aware Ordering Objective}
\label{sec:ordering_objective}

\noindent\textbf{Mismatch Approximation.}
Given the global conditional-TV matrix $W$, we
construct an approximation to the draft--target
mismatch within a speculative window.
When verifying the \(j\)-th draft, the target distribution incorporates preceding accepted drafts, whereas the draft distribution does not. 
Their joint influence is difficult to estimate directly.
We approximate this joint influence by summing the pairwise conditional-TV values:
\begin{equation}
\widehat{\delta}_{t,j}(\pi)
=
\sum_{i=1}^{j-1}
W_{\pi_{t+i}\rightarrow\pi_{t+j}},
\end{equation}
where \(t\) denotes the number of previously generated tokens, and \(j\) indexes the draft tokens within the speculative window.
This pairwise additive approximation provides a
computationally efficient mismatch score without
explicitly modeling higher-order interactions.

\noindent\textbf{Rejection-Cost Weighting.}
Motivated by Eq.~\ref{eq:expected_length_tv},
which assigns greater costs to earlier rejections,
we define a rejection-cost-aware window objective:
\begin{equation}
\mathcal{J}_K(\pi,t)
=
\sum_{j=1}^{k_t-1}
(k_t-j)\widehat{\delta}_{t,j}(\pi),
\end{equation}
where $k_t=\min(K,N-t)$ accounts for truncated
windows near the end of generation.
This objective combines the estimated mismatch
with position-dependent rejection cost, without
explicitly modeling prefix survival probabilities.

\noindent\textbf{Global Ordering Objective.}
Since the number of committed tokens varies across speculative rounds, subsequent speculative windows can start at different positions.
Therefore, optimizing a fixed partition into non-overlapping windows may overlook windows encountered during actual decoding.

To account for this variability, we aggregate
the window-level objectives over all possible
starting positions:
\begin{equation}
\mathcal{F}_K(\pi)
=
\sum_{t=0}^{N-1}
\rho_t \mathcal{J}_K(\pi,t),
\end{equation}
where we use uniform weights \(\rho_t=1/N\) rather than estimating the order-dependent visitation probabilities of window starts.

We then formulate generation-order optimization as
\begin{equation}
\min_{\pi\in\mathcal{S}_N}
\mathcal{F}_K(\pi).
\label{eq:global_objective}
\end{equation}

\subsection{Global Order Construction}
\label{sec:order_construction}

Exhaustively searching the $N!$ possible generation
orders is computationally infeasible.
However, the pairwise additive structure of
$\mathcal{F}_K$ allows us to efficiently construct
a generation order using the method of conditional
expectations~\cite{motwani1995randomized}.

\noindent\textbf{Pairwise Decomposition.}
The global objective can be rewritten as
\begin{equation}
\mathcal{F}_K(\pi)
=
\sum_{1\le r<s\le N}
C_{r,s}^{(K)}
W_{\pi_r\rightarrow\pi_s},
\end{equation}
where $C_{r,s}^{(K)}$ depends only on the
speculative window structure and the weights
in the global objective.

\noindent\textbf{Conditional-Expectation Ordering.}
Given a fixed prefix $\pi_{<m}$, let
$\mathcal{U}_m$ denote the remaining token positions.
For each candidate $v\in\mathcal{U}_m$,
we consider the expected global objective over
uniformly random completions:
\begin{equation}
Q_m(v)
=
\mathbb{E}_{\omega}
\left[
\mathcal{F}_K(\pi_{<m},v,\omega)
\right],
\end{equation}
where $\omega$ is uniformly distributed over
permutations of $\mathcal{U}_m\setminus\{v\}$.
Thanks to the pairwise additive structure,
$Q_m(v)$ can be computed analytically from $W$
without enumerating the remaining permutations.

We select the candidate minimizing this expectation:
\begin{equation}
\pi_m^{\mathrm{SAO}}
=
\arg\min_{v\in\mathcal{U}_m}Q_m(v).
\end{equation}
Repeating this procedure yields a complete
generation order $\pi_{\mathrm{SAO}}$.

With exact conditional expectations, the resulting
order satisfies
\begin{equation}
\mathcal{F}_K(\pi_{\mathrm{SAO}})
\leq
\mathbb{E}_{\pi\sim\operatorname{Unif}(\mathcal{S}_N)}
[\mathcal{F}_K(\pi)].
\end{equation}

The entire optimization is performed offline.
During inference, the constructed order is directly
used for self-speculative decoding, requiring no
additional model training or online order search.

  \begin{table*}[t]
    \centering
    \caption{Main results at $K=16$. Latency is measured in seconds per
    image with batch size one. NFE and latency speedups are relative to
    sequential decoding under the same order. Random-order results are
    reported as mean and standard deviation over five independently sampled generation orders. The best results among sequential and exact
    self-speculative methods are shown in \textbf{bold}; Direct Parallel serves as a quality--efficiency reference. \textcolor{red}{Red} highlights the speedups of Self-Spec SAO.}
    \label{tab:main}
    \resizebox{\textwidth}{!}{%
    \begin{tabular}{cl*{7}{c}}
        \toprule
        \textbf{Backbone} & \textbf{Method} & \textbf{NFE $\downarrow$} & $\mathbf{E}[L]\mathbf{\uparrow}$ & \textbf{Latency (s) $\downarrow$} & \multicolumn{2}{c}{\textbf{Speedup $\uparrow$}} & \textbf{FID $\downarrow$} & \textbf{IS $\uparrow$} \\
        \cmidrule(lr){6-7}
         & & & & & \textbf{NFE} & \textbf{Latency} & & \\
        \midrule
        \multirow{7}{*}{ARPG-L} & Sequential Raster & 256.00 & 1.00 & 4.82 & $1.00\times$ & $1.00\times$ & 40.82 & 33.99 \\
        & Sequential Random & 256.00 & 1.00 & $4.82{\scriptstyle\,\pm\,0.01}$ & $1.00\times$ & $1.00\times$ & $16.89{\scriptstyle\,\pm\,0.14}$ & $79.74{\scriptstyle\,\pm\,0.79}$ \\
         & Sequential SAO & 256.00 & 1.00 & 4.78 & $1.00\times$ & $1.00\times$ & \textbf{14.28} & \textbf{90.65} \\
        \addlinespace[5pt]
         & Self-Spec Raster & 123.32 & 4.14 & 2.39 & $2.08\times$ & $2.02\times$ & 41.00 & 33.88 \\
         & Self-Spec Random & $56.46{\scriptstyle\,\pm\,0.02}$ & $9.05{\scriptstyle\,\pm\,0.00}$ & $1.12{\scriptstyle\,\pm\,0.03}$ & $4.53{\scriptstyle\,\pm\,0.00}\times$ & $4.30{\scriptstyle\,\pm\,0.11}\times$ & $16.99{\scriptstyle\,\pm\,0.01}$ & $79.83{\scriptstyle\,\pm\,0.40}$ \\
         & \textbf{Self-Spec SAO} & \textbf{45.60} & \textbf{11.21} & \textbf{0.90} & \textcolor{red}{$\mathbf{5.61}\times$} & \textcolor{red}{$\mathbf{5.29}\times$} & 14.67 & 89.22 \\
        \cmidrule(lr){2-9}
         & Direct Parallel & 16.00 & 16.00 & $0.33{\scriptstyle\,\pm\,0.00}$ & $16.00\times$ & $14.40{\scriptstyle\,\pm\,0.02}\times$ & $38.28{\scriptstyle\,\pm\,0.19}$ & $44.70{\scriptstyle\,\pm\,0.38}$ \\
        \midrule
        \textbf{Backbone} & \textbf{Method} & \textbf{NFE $\downarrow$} & $\mathbf{E}[L]\mathbf{\uparrow}$ & \textbf{Latency (s) $\downarrow$} & \multicolumn{2}{c}{\textbf{Speedup $\uparrow$}} & \textbf{FID $\downarrow$} & \textbf{CLIP-Score $\uparrow$} \\
        \cmidrule(lr){6-7}
         & & & & & \textbf{NFE} & \textbf{Latency} & & \\
        \midrule
        \multirow{7}{*}{RLlamaGen} & Sequential Raster & 257.00 & 1.00 & 10.58 & $1.00\times$ & $1.00\times$ & 34.67 & 26.51 \\
        & Sequential Random & 257.00 & 1.00 & $10.52{\scriptstyle\,\pm\,0.01}$ & $1.00\times$ & $1.00\times$ & $31.03{\scriptstyle\,\pm\,0.05}$ & $27.34{\scriptstyle\,\pm\,0.01}$ \\
         & Sequential SAO & 257.00 & 1.00 & 10.21 & $1.00\times$ & $1.00\times$ & 30.40 & 27.39 \\
        \addlinespace[5pt]
        & Self-Spec Raster & 128.60 & 4.00 & 5.43 & $2.00\times$ & $1.95\times$ & 34.57 & 26.49 \\
         & Self-Spec Random & $64.34{\scriptstyle\,\pm\,0.09}$ & $8.07{\scriptstyle\,\pm\,0.01}$ & $2.73{\scriptstyle\,\pm\,0.03}$ & $3.99{\scriptstyle\,\pm\,0.01}\times$ & $3.85{\scriptstyle\,\pm\,0.04}\times$ & $30.93{\scriptstyle\,\pm\,0.21}$ & $27.34{\scriptstyle\,\pm\,0.00}$ \\
         & \textbf{Self-Spec SAO} & \textbf{55.54} & \textbf{9.37} & \textbf{2.25} & \textcolor{red}{$\mathbf{4.63}\times$} & \textcolor{red}{$\mathbf{4.53}\times$} & \textbf{30.33} & \textbf{27.42} \\
        \cmidrule(lr){2-9}
         & Direct Parallel & 33.00 & 16.00 & $1.39{\scriptstyle\,\pm\,0.00}$ & $7.79\times$ & $7.55{\scriptstyle\,\pm\,0.02}\times$ & $43.95{\scriptstyle\,\pm\,0.16}$ & $26.33{\scriptstyle\,\pm\,0.03}$ \\
        \bottomrule
    \end{tabular}%
    }
\end{table*}

\section{Experiments}
\label{sec:experiments}

\subsection{Experimental Setup}
\noindent\textbf{Implementation Details.}
We evaluate SAO on two random-order autoregressive models: ARPG-L~\cite{li2026arpg} for class-conditional generation on ImageNet-256~\cite{deng2009imagenet} and RLlamaGen~\cite{wu2025nep} for text-conditional generation on COCO val2014~\cite{lin2014coco}.
Quality evaluation uses 50K ImageNet and 30K COCO samples at $256\times256$ resolution. 
All model parameters are frozen. 
For each backbone, we construct a conditional-TV matrix from 5K offline profiling trajectories. 
For a fair comparison, both models use a classifier-free guidance scale of 1.0, temperature 1.0, top-$k=0$, and top-$p=1.0$.
Detailed settings are provided in the appendix.

\noindent\textbf{Metrics.}
We evaluate image quality using FID~\cite{heusel2017fid}, and additionally report Inception Score (IS)~\cite{salimans2016improved} for class-conditional generation and CLIP~\cite{radford2021clip} cosine similarity multiplied by 100 for text-conditional generation.
For decoding efficiency, we report the number of function evaluations (NFE) per image and the average number of committed tokens per speculative round, $\mathbb{E}[L]$.
We measure latency with batch size one on an NVIDIA A100 40GB GPU.
We report NFE and latency speedups relative to sequential decoding under
the same order, and order speedup as the ratio of Random-order latency to SAO-order latency for the same decoder.
Further details are provided in the appendix.

\noindent\textbf{Baselines.}
We compare sequential decoding under Raster, Random, and SAO orders (Sequential Raster, Sequential Random, and Sequential SAO), exact self-speculative decoding under the same three ordering strategies (Self-Spec Raster, Self-Spec Random, and Self-Spec SAO), and direct parallel decoding under a random order (Direct Parallel).
For random-order methods, we report the mean and standard deviation over independently sampled generation orders.

\begin{figure*}[t]
      \centering
      \includegraphics[width=\textwidth]{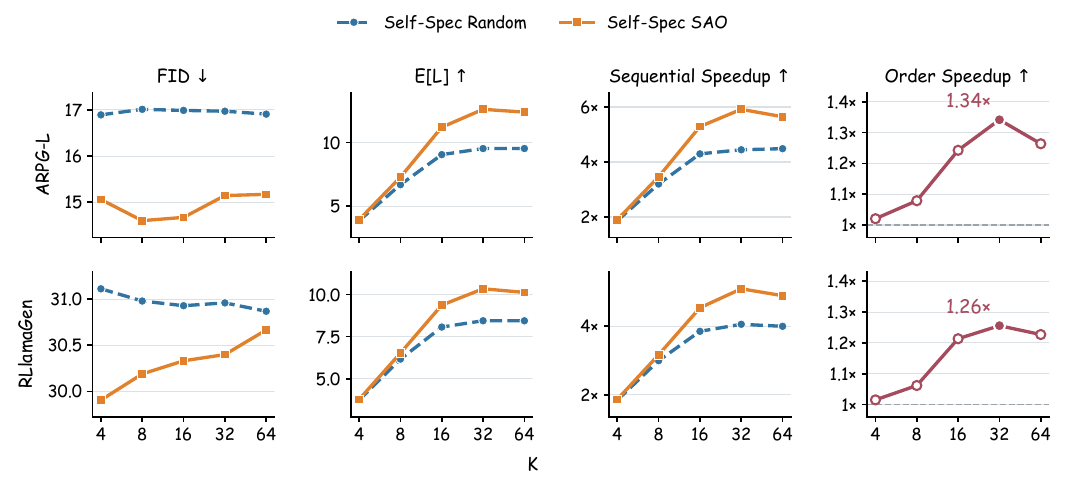}
      \caption{Effect of speculative width $K$ on FID, average committed length $\mathbb{E}[L]$, sequential speedup, and order speedup.
      The top and bottom rows show ARPG-L and RLlamaGen, respectively. 
      Sequential speedup is measured relative to sequential decoding under the same order. 
      Order speedup is the ratio of mean Self-Spec Random latency to Self-Spec SAO latency. Random-order curves report means over five independently sampled generation orders.}
      \label{fig:k_scaling}
\end{figure*}

\begin{figure}[t]
      \centering
      \includegraphics[width=\columnwidth]{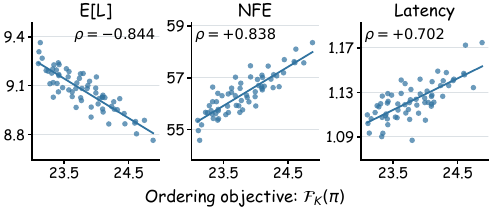}
      \caption{Ordering objective versus measured decoding efficiency. Each point represents one random generation order. Lines show linear fits, and $\rho$ denotes Spearman's rank correlation.}
      \label{fig:ordering_objective}
\end{figure}

\subsection{Main Results}

\noindent\textbf{Self-speculative decoding accelerates generation.}
Our exact self-speculative decoder reduces NFE and latency relative to sequential decoding under Raster, Random, and SAO orders on both backbones.
Under the SAO order, it achieves latency speedups of $5.29\times$ on ARPG-L and $4.53\times$ on RLlamaGen relative to same-order sequential decoding.

\noindent\textbf{SAO further improves decoding efficiency.}
As shown in \cref{tab:main}, simply switching from Raster to Random order reduces NFE from 123.32 to 56.46 on ARPG-L and from 128.60 to 64.34 on RLlamaGen, using the same model and self-speculative decoder.
These gains highlight the importance of generation order in optimizing speculative decoding.
SAO further improves efficiency through explicit order optimization.
On ARPG-L, SAO increases the average number of committed tokens per round from 9.05 to 11.21, reducing the NFE from 56.46 to 45.60 per image.
The resulting latency decreases from 1.12\,s to 0.90\,s, yielding an order speedup of $1.24\times$.
On RLlamaGen, SAO similarly increases the average committed length from 8.07 to 9.37 and reduces the NFE from 64.34 to 55.54.
The latency decreases from 2.73\,s to 2.25\,s, corresponding to an order speedup of $1.21\times$.

\noindent\textbf{SAO also improves generation quality.}
We first compare sequential and self-speculative decoding under the same SAO order.
As shown in \cref{tab:main}, they obtain similar FID scores on ARPG-L (14.28 vs.\ 14.67) and RLlamaGen (30.40 vs.\ 30.33).
On RLlamaGen, their CLIP scores are also close (27.39 vs.\ 27.42).
This observation is consistent with the distribution-preserving property of exact self-speculative decoding under a fixed order.

We next examine the effect of generation order.
Sequential SAO achieves lower FID than Sequential Random on both ARPG-L (14.28 vs.\ 16.89) and RLlamaGen (30.40 vs.\ 31.03).
The same trend is observed with self-speculative decoding: Self-Spec SAO achieves lower FID than Self-Spec Random on both backbones.
These results indicate that the observed quality improvement is associated with the generation order and is not limited to self-speculative decoding.

One possible explanation is that a finite-capacity model may approximate the conditional distributions induced by different orders with different levels of accuracy.
SAO favors orders with lower estimated conditional influence within speculative windows.
This may make some predictions less sensitive to preceding token values and potentially easier for the model to approximate.

Compared with Self-Spec SAO, Direct Parallel achieves lower latency on both backbones but suffers from substantial degradation in generation quality.
In contrast, Self-Spec SAO preserves exact sampling under the chosen generation order while improving decoding efficiency, offering a more balanced efficiency--quality trade-off.

\begin{table*}[t]
    \centering
    \caption{Complementarity of SAO with existing self-speculative decoding methods on both backbones at $K=16$.  Latency is measured in seconds per image at batch size one. Random-order results are reported as mean and standard deviation over five independently sampled generation orders. Order speedups are relative to random ordering under the same decoding method. \textcolor{red}{Red} highlights the order speedups achieved by SAO.}
    \label{tab:complementarity}
    \setlength{\tabcolsep}{5pt}
    \begin{tabular}{lllccccc}
        \toprule
        \textbf{Backbone} & \textbf{Decoder} & \textbf{Order} & \textbf{NFE} $\downarrow$ & \textbf{Latency (s)} $\downarrow$ & \multicolumn{2}{c}{\textbf{Latency Speedup} $\uparrow$} & \textbf{FID} $\downarrow$ \\
        \cmidrule(lr){6-7}
        & & & & & \textbf{Sequential} & \textbf{Order} & \\
        \midrule
        ARPG-L & Self-Spec & Random & $56.46{\scriptstyle\,\pm\,0.02}$ & $1.12{\scriptstyle\,\pm\,0.03}$ & $4.30{\scriptstyle\,\pm\,0.11}\times$ & -- & $16.99{\scriptstyle\,\pm\,0.01}$ \\
         &  & \textbf{SAO} & 45.60 & 0.90 & $5.29\times$ & \textcolor{red}{$\mathbf{1.24}\times$} & 14.67 \\
        \addlinespace[2pt]
         & SJD & Random & $43.15{\scriptstyle\,\pm\,0.00}$ & $1.40{\scriptstyle\,\pm\,0.02}$ & $3.44{\scriptstyle\,\pm\,0.04}\times$ & -- & $16.93{\scriptstyle\,\pm\,0.08}$ \\
         &  & \textbf{SAO} & 37.12 & 1.21 & $3.94\times$ & \textcolor{red}{$\mathbf{1.16}\times$} & 14.55 \\
        \addlinespace[2pt]
         & SCD-MC & Random & $40.23{\scriptstyle\,\pm\,0.00}$ & $1.26{\scriptstyle\,\pm\,0.01}$ & $3.83{\scriptstyle\,\pm\,0.01}\times$ & -- & $16.89{\scriptstyle\,\pm\,0.10}$ \\
         &  & \textbf{SAO} & 36.38 & 1.13 & $4.24\times$ & \textcolor{red}{$\mathbf{1.12}\times$} & 14.52 \\
        \addlinespace[2pt]
         & SCD-GS & Random & $40.34{\scriptstyle\,\pm\,0.01}$ & $1.32{\scriptstyle\,\pm\,0.00}$ & $3.66{\scriptstyle\,\pm\,0.02}\times$ & -- & $16.91{\scriptstyle\,\pm\,0.16}$ \\
         &  & \textbf{SAO} & 36.39 & 1.18 & $4.06\times$ & \textcolor{red}{$\mathbf{1.12}\times$} & 14.31 \\
        \midrule
        RLlamaGen & Self-Spec & Random & $64.34{\scriptstyle\,\pm\,0.09}$ & $2.73{\scriptstyle\,\pm\,0.03}$ & $3.85{\scriptstyle\,\pm\,0.04}\times$ & -- & $30.93{\scriptstyle\,\pm\,0.21}$ \\
         &  & \textbf{SAO} & 55.54 & 2.25 & $4.53\times$ & \textcolor{red}{$\mathbf{1.21}\times$} & 30.33 \\
        \addlinespace[2pt]
         & SJD & Random & $45.39{\scriptstyle\,\pm\,0.04}$ & $2.03{\scriptstyle\,\pm\,0.03}$ & $5.17{\scriptstyle\,\pm\,0.07}\times$ & -- & $31.00{\scriptstyle\,\pm\,0.09}$ \\
         &  & \textbf{SAO} & 39.75 & 1.73 & $5.90\times$ & \textcolor{red}{$\mathbf{1.17}\times$} & 30.39 \\
        \addlinespace[2pt]
         & SCD-MC & Random & $41.84{\scriptstyle\,\pm\,0.00}$ & $1.82{\scriptstyle\,\pm\,0.01}$ & $5.78{\scriptstyle\,\pm\,0.04}\times$ & -- & $30.92{\scriptstyle\,\pm\,0.02}$ \\
         &  & \textbf{SAO} & 38.48 & 1.62 & $6.30\times$ & \textcolor{red}{$\mathbf{1.12}\times$} & 30.19 \\
        \addlinespace[2pt]
         & SCD-GS & Random & $41.97{\scriptstyle\,\pm\,0.02}$ & $1.87{\scriptstyle\,\pm\,0.01}$ & $5.62{\scriptstyle\,\pm\,0.03}\times$ & -- & $30.71{\scriptstyle\,\pm\,0.12}$ \\
         &  & \textbf{SAO} & 38.52 & 1.68 & $6.08\times$ & \textcolor{red}{$\mathbf{1.11}\times$} & 30.57 \\
        \bottomrule
    \end{tabular}
\end{table*}

\noindent\textbf{SAO remains effective across speculative widths.}
The speculative width \(K\) determines the maximum number of tokens that can be committed in each decoding round.
A larger \(K\) allows more tokens to be committed per round, but early rejections and verification overhead may limit the resulting speedup.
We therefore evaluate SAO across different speculative widths.

As shown in \cref{fig:k_scaling}, Self-Spec SAO commits more tokens per round than Self-Spec Random on both backbones across all evaluated speculative widths. 
At $K=32$, $\mathbb{E}[L]$ increases from 9.53 to 12.62 on ARPG-L and from 8.45 to 10.34 on RLlamaGen.
The rightmost column shows that order speedup is close to $1.0\times$ at $K=4$ and peaks at $K=32$ on both backbones: $1.34\times$ on ARPG-L and $1.26\times$ on RLlamaGen.

Self-Spec SAO also achieves lower FID than Self-Spec Random across all evaluated speculative widths on both backbones.

\subsection{Validating the ordering objective}

We test whether the global ordering objective $\mathcal{F}_K(\pi)$ in \cref{eq:global_objective} predicts which generation orders lead to more efficient self-speculative decoding.
We evaluate 64 uniformly sampled generation orders on ARPG-L at $K=16$,
keeping the decoder and sampling settings fixed.
As shown in  \cref{fig:ordering_objective}, lower objective values correspond to more tokens committed per round, fewer function evaluations per image, and lower latency.
These correlations suggest that $\mathcal{F}_K(\pi)$ can serve as an
offline proxy for ranking generation orders by decoding efficiency.

\subsection{Complementarity with Other Self-Speculative Decoding Methods}

The results above demonstrate the effectiveness of generation order as an additional degree of freedom for improving self-speculative decoding.
We next investigate the complementarity between generation order optimization and existing self-speculative decoding methods.
We evaluate SJD~\cite{teng2025sjd} and two SCD variants, SCD-MC and SCD-GS~\cite{so2026scd}, under Random and SAO orders with the same sampling settings at $K=16$.

As shown in \cref{tab:complementarity}, SAO reduces both NFE and latency for all evaluated decoders on both backbones.
Across both backbones, SAO achieves order speedups of $1.16$--$1.17\times$ for SJD and $1.11$--$1.12\times$ for the two SCD variants.
These results suggest that generation-order optimization complements different self-speculative drafting methods.
Beyond improving decoding efficiency, SAO also improves generation quality as measured by FID, yielding lower FID than Random order for every evaluated decoder on both backbones.

Interestingly, the relative latency of different decoders varies across backbones.
On ARPG-L, our self-speculative decoder achieves lower latency than SJD and SCD despite requiring more function evaluations, whereas the opposite trend is observed on RLlamaGen.
This difference may be explained by their
architectural designs.
ARPG decouples content representations from position-guided queries, enabling efficient parallel drafting through a shared KV cache.
In contrast, RLlamaGen follows a conventional causal Transformer architecture with shifted positional embeddings, where parallel drafting
must be implemented through the standard
decoder computation.
These results suggest that practical decoding efficiency depends not only on NFE but also on how efficiently the underlying architecture supports parallel token prediction.

\subsection{Ablation Study}

\begin{table}[t]
    \centering
    \caption{Ablation of SAO on both backbones at $K=16$.
    Random and Shuffled $W$ results are reported as mean and standard deviation over five random orders and five matrix permutations, respectively.}
    \label{tab:ablation}
    \setlength{\tabcolsep}{3pt}
    \resizebox{\columnwidth}{!}{%
    \begin{tabular}{lcccc}
        \toprule
        \textbf{Order} & $\mathbf{E}[L]\boldsymbol{\uparrow}$ & \textbf{NFE} $\downarrow$ & \textbf{Latency (s)} $\downarrow$ & \textbf{FID} $\downarrow$ \\
        \midrule
        \multicolumn{5}{l}{\textit{ARPG-L}} \\
        Random & $9.05{\scriptstyle\,\pm\,0.00}$ & $56.46{\scriptstyle\,\pm\,0.02}$ & $1.12{\scriptstyle\,\pm\,0.03}$ & $16.99{\scriptstyle\,\pm\,0.01}$ \\
        Shuffled $W$ & $9.09{\scriptstyle\,\pm\,0.08}$ & $56.16{\scriptstyle\,\pm\,0.43}$ & $1.13{\scriptstyle\,\pm\,0.02}$ & $16.52{\scriptstyle\,\pm\,0.28}$ \\
        Unweighted & 10.80 & 47.42 & 0.99 & 14.80 \\
        Blockwise & 10.18 & 50.05 & 1.02 & 15.13 \\
        \textbf{SAO} & \textbf{11.21} & \textbf{45.60} & \textbf{0.90} & \textbf{14.67} \\
        \midrule
        \multicolumn{5}{l}{\textit{RLlamaGen}} \\
        Random & $8.07{\scriptstyle\,\pm\,0.01}$ & $64.34{\scriptstyle\,\pm\,0.09}$ & $2.73{\scriptstyle\,\pm\,0.03}$ & $30.93{\scriptstyle\,\pm\,0.21}$ \\
        Shuffled $W$ & $7.97{\scriptstyle\,\pm\,0.07}$ & $65.07{\scriptstyle\,\pm\,0.47}$ & $2.79{\scriptstyle\,\pm\,0.02}$ & $30.84{\scriptstyle\,\pm\,0.26}$ \\
        Unweighted & 9.03 & 57.68 & 2.34 & 30.51 \\
        Blockwise & 8.76 & 59.26 & 2.53 & 30.60 \\
        \textbf{SAO} & \textbf{9.37} & \textbf{55.54} & \textbf{2.25} & \textbf{30.33} \\
        \bottomrule
    \end{tabular}%
    }
\end{table}

We ablate three design choices in SAO:
the conditional-TV proxy, rejection-cost weighting, and global order construction.
All variants use the same pretrained generator, self-speculative decoder, and sampling configuration, differing only in how the generation order is constructed.
We consider the following variants:

\begin{itemize}
      \item \textbf{Shuffled $W$.}
      We replace the original influence matrix with
      $\widetilde{W}_{uv}=W_{\tau(u),\tau(v)}$, where $\tau$ is a
      permutation of visual token positions.
      This preserves the matrix values while disrupting their
      correspondence to the original positions.

      \item \textbf{Unweighted.}
      We replace each rejection-cost weight \((k_t-j)\) in the window-level objective with \(1\), while preserving the conditional-TV matrix and global aggregation over window starts.

      \item \textbf{Blockwise.}
      We construct the generation order in consecutive, non-overlapping groups of size $K$. Within each group, positions are selected sequentially from the remaining tokens using the method of conditional expectations, minimizing the expected weighted objective of the current group under random completion. 
      Unlike SAO, Blockwise optimizes only these fixed groups and ignores windows crossing group boundaries. 
\end{itemize}
We additionally include Random order as a baseline.

As shown in \cref{tab:ablation}, Shuffled $W$ performs similarly to random ordering on both backbones, indicating that meaningful pairwise conditional influence is essential for effective order optimization.

Unweighted improves upon random ordering on both backbones but remains inferior to SAO, suggesting that considering draft--target mismatch alone is insufficient without accounting for early-rejection cost.

Compared with blockwise ordering, SAO increases the average committed length from 10.18 to 11.21 on ARPG-L and from 8.76 to 9.37 on RLlamaGen.
Correspondingly, NFE decreases from 50.05 to 45.60 and from 59.26 to 55.54, respectively.
These results highlight the benefit of accounting for varying speculative window starts rather than optimizing fixed, non-overlapping windows.

  \section{Related Work}
\label{sec:related_work}

\noindent\textbf{Random-Order Visual Autoregressive Generation.}
Traditional autoregressive image generation typically follows a fixed raster-scan order~\cite{esser2021taming,sun2024autoregressive}.
Recent works have explored random-order modeling to overcome this constraint.
RAR~\cite{yu2025rar} introduces randomized factorization orders during training to improve visual context modeling.
RandAR~\cite{pang2025randar} enables arbitrary-order generation through position instruction tokens, while ARPG~\cite{li2026arpg} decouples positional guidance from content representations to support efficient parallel prediction with a shared KV cache.
LPD~\cite{zhang2026locality} further combines flexible parallel generation with locality-aware ordering.
RLlamaGen~\cite{wu2025nep} extends LlamaGen to arbitrary-order text-to-image generation through shifted positional embeddings.
These advances provide a foundation for self-speculative decoding: efficient parallel prediction allows the generator itself to serve as a fast draft model, while flexible generation orders offer an additional dimension for optimizing decoding efficiency.

\noindent\textbf{Speculative Decoding for Visual Generation.}
Speculative decoding accelerates autoregressive generation by drafting multiple tokens and verifying them against the target model~\cite{leviathan2023speculative,chen2023speculative}.
For visual generation, LANTERN~\cite{jang2025lantern} relaxes the acceptance criterion to address visual token ambiguity, while GSD~\cite{so2025gsd} improves acceptance through dynamic token grouping.
LISA~\cite{li2026lisa} further exploits visual locality to improve drafting and verification.

To eliminate the need for a separate draft model, self-speculative decoding leverages parallel generation for drafting and sequential generation for verification within the same model.
SJD~\cite{teng2025sjd} introduces training-free speculative Jacobi decoding, while SCD~\cite{so2026scd} improves draft consistency through coupling.
SJD-PAC~\cite{kang2026sjdpac} further enhances drafting and enables verification to continue after rejection.
While these methods primarily focus on drafting, acceptance, or verification, SAO explores a complementary dimension: optimizing generation order to improve self-speculative decoding efficiency.

  \section{Conclusion}

We are the first to investigate generation order as an optimization dimension for self-speculative
decoding in autoregressive image generation.
We theoretically analyze how generation order affects decoding efficiency through draft--target mismatch and early-rejection cost.
Building on these insights, we propose Speculation-Aware Ordering (SAO), a training-free framework that constructs speculation-friendly generation orders entirely offline.
Experiments demonstrate consistent efficiency improvements across different backbones and speculative widths, with additional gains in
generation quality and complementary acceleration when combined with existing self-speculative decoding methods.

  {\small
  \bibliographystyle{ieeenat_fullname}
  \bibliography{main}

@inproceedings{brown2020language,
  title     = {Language Models are Few-Shot Learners},
  author    = {Brown, Tom and Mann, Benjamin and Ryder, Nick and Subbiah, Melanie and Kaplan, Jared D. and Dhariwal, Prafulla and Neelakantan, Arvind and Shyam, Pranav and Sastry, Girish and Askell, Amanda and Agarwal, Sandhini and Herbert-Voss, Ariel and Krueger, Gretchen and Henighan, Tom and Child, Rewon and Ramesh, Aditya and Ziegler, Daniel and Wu, Jeffrey and Winter, Clemens and Hesse, Chris and Chen, Mark and Sigler, Eric and Litwin, Mateusz and Gray, Scott and Chess, Benjamin and Clark, Jack and Berner, Christopher and McCandlish, Sam and Radford, Alec and Sutskever, Ilya and Amodei, Dario},
  booktitle = {Advances in Neural Information Processing Systems},
  volume    = {33},
  year      = {2020}
}

@inproceedings{esser2021taming,
  title     = {Taming Transformers for High-Resolution Image Synthesis},
  author    = {Esser, Patrick and Rombach, Robin and Ommer, Bj{\"o}rn},
  booktitle = {Proceedings of the IEEE/CVF Conference on Computer Vision and Pattern Recognition},
  pages     = {12873--12883},
  year      = {2021}
}

@article{sun2024autoregressive,
  title   = {Autoregressive Model Beats Diffusion: {Llama} for Scalable Image Generation},
  author  = {Sun, Peize and Jiang, Yi and Chen, Shoufa and Zhang, Shilong and Peng, Bingyue and Luo, Ping and Yuan, Zehuan},
  journal = {arXiv preprint arXiv:2406.06525},
  year    = {2024}
}

@inproceedings{pang2025randar,
  title     = {{RandAR}: Decoder-only Autoregressive Visual Generation in Random Orders},
  author    = {Pang, Ziqi and Zhang, Tianyuan and Luan, Fujun and Man, Yunze and Tan, Hao and Zhang, Kai and Freeman, William T. and Wang, Yu-Xiong},
  booktitle = {Proceedings of the IEEE/CVF Conference on Computer Vision and Pattern Recognition},
  pages     = {45--55},
  year      = {2025}
}

@inproceedings{li2026arpg,
  title     = {Autoregressive Image Generation with Randomized Parallel Decoding},
  author    = {Li, Haopeng and Yang, Jinyue and Li, Guoqi and Wang, Huan},
  booktitle = {International Conference on Learning Representations},
  year      = {2026}
}

@misc{radford2019language,
  title        = {Language Models are Unsupervised Multitask Learners},
  author       = {Radford, Alec and Wu, Jeffrey and Child, Rewon and Luan, David and Amodei, Dario and Sutskever, Ilya},
  year         = {2019},
  howpublished = {OpenAI technical report},
  url          = {https://cdn.openai.com/better-language-models/language_models_are_unsupervised_multitask_learners.pdf}
}

@inproceedings{leviathan2023speculative,
  title = {Fast Inference from Transformers via Speculative Decoding},
  author = {Leviathan, Yaniv and Kalman, Matan and Matias, Yossi},
  booktitle = {Proceedings of the 40th International Conference on Machine Learning},
  pages = {19274--19286},
  year = {2023}
}

@article{chen2023speculative,
  title = {Accelerating Large Language Model Decoding with Speculative Sampling},
  author = {Chen, Charlie and Borgeaud, Sebastian and Irving, Geoffrey and Lespiau, Jean-Baptiste and Sifre, Laurent and Jumper, John},
  journal = {arXiv preprint arXiv:2302.01318},
  year = {2023}
}

@inproceedings{teng2025sjd,
  title = {Accelerating Auto-regressive Text-to-Image Generation with Training-free Speculative Jacobi Decoding},
  author = {Teng, Yao and Shi, Han and Liu, Xian and Ning, Xuefei and Dai, Guohao and Wang, Yu and Li, Zhenguo and Liu, Xihui},
  booktitle = {International Conference on Learning Representations},
  year = {2025}
}

@inproceedings{so2026scd,
  title = {Speculative Coupled Decoding for Training-Free Lossless Acceleration of Autoregressive Visual Generation},
  author = {So, Junhyuk and Kook, Hyunho and Jang, Chaeyeon and Park, Eunhyeok},
  booktitle = {Proceedings of the 43rd International Conference on Machine Learning},
  pages = {114051--114074},
  year = {2026}
}

@inproceedings{guo2026assd,
  title = {Self-Speculative Decoding Accelerates Lossless Inference in Any-Order and Any-Subset Autoregressive Models},
  author = {Guo, Gabe and Ermon, Stefano},
  booktitle = {International Conference on Learning Representations},
  year = {2026}
}

@inproceedings{zhang2026locality,
  title = {Locality-aware Parallel Decoding for Efficient Autoregressive Image Generation},
  author = {Zhang, Zhuoyang and Huang, Luke and Wu, Chengyue and Yang, Shang and Peng, Kelly and Lu, Yao and Han, Song},
  booktitle = {International Conference on Learning Representations},
  year = {2026}
}

@inproceedings{so2025gsd,
  title = {Grouped Speculative Decoding for Autoregressive Image Generation},
  author = {So, Junhyuk and Shin, Juncheol and Kook, Hyunho and Park, Eunhyeok},
  booktitle = {Proceedings of the IEEE/CVF International Conference on Computer Vision},
  pages = {15375--15384},
  year = {2025}
}

@book{motwani1995randomized,
  title     = {Randomized Algorithms},
  author    = {Motwani, Rajeev and Raghavan, Prabhakar},
  publisher = {Cambridge University Press},
  year      = {1995}
}

@inproceedings{wu2025nep,
  title = {{NEP}: Autoregressive Image Editing via Next Editing Token Prediction},
  author = {Wu, Huimin and Ma, Xiaojian and Zhao, Haozhe and Zhao, Yanpeng and Li, Qing},
  booktitle = {Advances in Neural Information Processing Systems},
  volume = {38},
  year = {2025}
}

@inproceedings{deng2009imagenet,
  title = {{ImageNet}: A Large-Scale Hierarchical Image Database},
  author = {Deng, Jia and Dong, Wei and Socher, Richard and Li, Li-Jia and Li, Kai and Fei-Fei, Li},
  booktitle = {Proceedings of the IEEE Conference on Computer Vision and Pattern Recognition},
  pages = {248--255},
  year = {2009}
}

@inproceedings{lin2014coco,
  title = {Microsoft {COCO}: Common Objects in Context},
  author = {Lin, Tsung-Yi and Maire, Michael and Belongie, Serge and Hays, James and Perona, Pietro and Ramanan, Deva and Doll{\'a}r, Piotr and Zitnick, C. Lawrence},
  booktitle = {European Conference on Computer Vision},
  pages = {740--755},
  year = {2014}
}

@inproceedings{heusel2017fid,
  title = {{GANs} Trained by a Two Time-Scale Update Rule Converge to a Local {Nash} Equilibrium},
  author = {Heusel, Martin and Ramsauer, Hubert and Unterthiner, Thomas and Nessler, Bernhard and Hochreiter, Sepp},
  booktitle = {Advances in Neural Information Processing Systems},
  volume = {30},
  year = {2017}
}

@inproceedings{salimans2016improved,
  title = {Improved Techniques for Training {GANs}},
  author = {Salimans, Tim and Goodfellow, Ian and Zaremba, Wojciech and Cheung, Vicki and Radford, Alec and Chen, Xi},
  booktitle = {Advances in Neural Information Processing Systems},
  volume = {29},
  year = {2016}
}

@inproceedings{radford2021clip,
  title = {Learning Transferable Visual Models From Natural Language Supervision},
  author = {Radford, Alec and Kim, Jong Wook and Hallacy, Chris and Ramesh, Aditya and Goh, Gabriel and Agarwal, Sandhini and Sastry, Girish and Askell, Amanda and Mishkin, Pamela and Clark, Jack and Krueger, Gretchen and Sutskever, Ilya},
  booktitle = {Proceedings of the 38th International Conference on Machine Learning},
  pages = {8748--8763},
  year = {2021}
}

@inproceedings{yu2025rar,
  title     = {Randomized Autoregressive Visual Generation},
  author    = {Yu, Qihang and He, Ju and Deng, Xueqing and Shen, Xiaohui and Chen, Liang-Chieh},
  booktitle = {Proceedings of the IEEE/CVF International Conference on Computer Vision},
  pages     = {18431--18441},
  year      = {2025}
}

@inproceedings{jang2025lantern,
  title = {{LANTERN}: Accelerating Visual Autoregressive Models with Relaxed Speculative Decoding},
  author = {Jang, Doohyuk and Park, Sihwan and Yang, June Yong and Jung, Yeonsung and Yun, Jihun and Kundu, Souvik and Kim, Sung-Yub and Yang, Eunho},
  booktitle = {International Conference on Learning Representations},
  year = {2025}
}

@inproceedings{li2026lisa,
  title = {{LISA}: Locality-Informed Speculative Decoding for Accelerating Autoregressive Image Generation},
  author = {Li, Ying and Jian, Siyong and Wang, Zhaode and Chen, Zhiwen and Lv, Chengfei and Wang, Huan},
  booktitle = {European Conference on Computer Vision},
  year = {2026}
}

@inproceedings{kang2026sjdpac,
  title = {{SJD-PAC}: Accelerating Speculative Jacobi Decoding via Proactive Drafting and Adaptive Continuation},
  author = {Kang, Jialiang and Shu, Han and Li, Wenshuo and Zhai, Yingjie and Chen, Xinghao},
  booktitle = {Proceedings of the IEEE/CVF Conference on Computer Vision and Pattern Recognition},
  year = {2026}
}

@inproceedings{wang2025par,
  author = {Wang, Yuqing and Ren, Shuhui and Lin, Zhijie and Han, Yujin and Guo, Haoyuan and Yang, Zhenheng and Zou, Difan and Feng, Jiashi and Liu, Xihui},
  title = {Parallelized Autoregressive Visual Generation},
  booktitle = {Proceedings of the IEEE/CVF Conference on Computer Vision and Pattern Recognition},
  pages = {12955--12965},
  year = {2025}
}

@article{halton1960efficiency,
  author = {Halton, John H.},
  title = {On the Efficiency of Certain Quasi-Random Sequences of Points in Evaluating Multi-Dimensional Integrals},
  journal = {Numerische Mathematik},
  volume = {2},
  pages = {84--90},
  year = {1960},
  doi = {10.1007/BF01386213}
}

@article{ho2022classifierfree,
  title         = {Classifier-Free Diffusion Guidance},
  author        = {Ho, Jonathan and Salimans, Tim},
  journal       = {arXiv preprint arXiv:2207.12598},
  year          = {2022}
}
  }

  \clearpage
\appendix

\section*{Appendix}

\section{Theoretical and Algorithmic Details}
\label{sec:suppl_theory}

\subsection{Derivation of the Pairwise Global Objective}
\label{sec:suppl_pairwise}

Let $\pi=(\pi_1,\ldots,\pi_N)$ be an order over the $N$ visual positions.
The conditional-TV proxy $W_{u\rightarrow v}$ measures how revealing position $u$ changes the prediction at position $v$:
\begin{align*}
d_{u,v}(h,a)
&=\operatorname{TV}\!\left(
p_\theta(x_v\mid h),\,
p_\theta(x_v\mid h\oplus(u,a))
\right),\\
W_{u\rightarrow v}
&=\mathbb{E}_{h\sim\mathcal D^{uv}}
  \mathbb{E}_{a\sim p_\theta(x_u\mid h)}
  [d_{u,v}(h,a)].
\end{align*}
Here $h$ is a profiling history drawn from $\mathcal D^{uv}$, in which neither position has been generated, and $h\oplus(u,a)$ appends the sampled value $a$ at $u$.

A window starting after $t$ committed tokens has width $k_t=\min(K,N-t)$.
For its $j$-th draft, SAO estimates mismatch by summing the influences of all preceding drafts and weights that estimate by the progress lost upon rejection:
\begin{align*}
\widehat{\delta}_{t,j}(\pi)
&=\sum_{i=1}^{j-1}W_{\pi_{t+i}\rightarrow\pi_{t+j}},\\
\mathcal J_K(\pi,t)
&=\sum_{j=1}^{k_t-1}(k_t-j)\widehat{\delta}_{t,j}(\pi).
\end{align*}
Averaging the window objectives uniformly and substituting the mismatch estimate gives
\begin{equation}
\mathcal{F}_K(\pi)
=\frac{1}{N}\sum_{t=0}^{N-1}
 \sum_{j=1}^{k_t-1}(k_t-j)
 \sum_{i=1}^{j-1}W_{\pi_{t+i}\rightarrow\pi_{t+j}}.
\end{equation}
Set $r=t+i$ and $s=t+j$.
The index constraints become $t<r<s<t+k_t$, and the rejection-cost weight becomes $k_t-s+t$.
Collecting the terms for each pair $1\le r<s\le N$ yields Eq.~(19) of the main paper:
\[
\mathcal F_K(\pi)
=\sum_{1\le r<s\le N}C_{r,s}^{(K)}W_{\pi_r\rightarrow\pi_s},
\]
where
\begin{equation}
C_{r,s}^{(K)}
=\frac{1}{N}\sum_{t=0}^{N-1}
   \mathbf{1}\{t<r<s<t+k_t\}\,(k_t-s+t).
\label{eq:suppl_ce_coeff}
\end{equation}
The strict upper bound on $s$ excludes the bonus position $t+k_t$.
At the end of generation, the same expression uses the actual remaining width $k_t$.
A window of width one or two contributes no pair because it has at most one draft position.

\subsection{Conditional-Expectation Order Construction}
\label{sec:suppl_ce_formula}

We use the method of conditional expectations~\cite{motwani1995randomized} to construct the generation order.

\noindent\textbf{Analytic candidate score.}
Write $C_{r,s}=C_{r,s}^{(K)}$ and $W_{u,v}=W_{u\rightarrow v}$.
Suppose candidate $v$ is tentatively appended at order index $m$, giving prefix $P'=(p_1,\ldots,p_m)$ with $p_m=v$.
Let $U'$ contain the $n$ positions outside $P'$.
The candidate score (Eq.~(20) of the main paper) is
\[
Q_m(v)=\mathbb{E}_{\omega}[\mathcal F_K(P'\oplus\omega)],
\]
where $\omega$ is a uniformly random permutation of $U'$, and $P'\oplus\omega$ denotes concatenation.
Under uniform completion, one future index is uniform over $U'$, and two distinct future indices are uniform over the $n(n-1)$ ordered pairs of distinct positions in $U'$.
Splitting the sum into fixed pairs, pairs with one future index, and pairs with two future indices gives
\begin{align}
Q_m(v)
={}&\sum_{1\le r<s\le m}C_{r,s}W_{p_r,p_s}
\notag\\
&+\frac{1}{n}\sum_{r=1}^{m}
  \left(\sum_{s>m}C_{r,s}\right)
  \sum_{u\in U'}W_{p_r,u}
\notag\\
&+\frac{\sum_{m<r<s\le N}C_{r,s}}{n(n-1)}
  \sum_{\substack{u,w\in U'\\u\ne w}}W_{u,w}.
\label{eq:suppl_ce_expectation}
\end{align}
The second term is omitted when $n=0$, and the third when $n<2$.

\noindent\textbf{Incremental scoring.}
Before selecting index $m$, let $P=(p_1,\ldots,p_{m-1})$ and let $U$ be the remaining set, so $n=|U|-1$.
With the unused diagonal of $W$ set to zero, maintain
\begin{align*}
R(a)&=\sum_{u\in U}W_{a,u}, &
I(a)&=\sum_{u\in U}W_{u,a},\\
D&=\sum_{a\in U}R(a), &
B&=\sum_{1\le r<s<m}C_{r,s}W_{p_r,p_s}.
\end{align*}
Precompute $H_{r,m}=\sum_{s>m}C_{r,s}$ and $T_m=\sum_{m<r<s\le N}C_{r,s}$.
For a trial candidate $v\in U$, define $\Delta_m(v)=\sum_{r<m}C_{r,m}W_{p_r,v}$.
Removing $v$ subtracts $W_{p_r,v}$ from each prefix row sum and subtracts $R(v)+I(v)$ from $D$.
The score can therefore be evaluated as
\begin{align}
Q_m(v)
={}&B+\Delta_m(v)
\notag\\
&+\frac{\sum_{r<m}H_{r,m}(R(p_r)-W_{p_r,v})}{n}
\notag\\
&+\frac{H_{m,m}R(v)}{n}
\notag\\
&+\frac{T_m}{n(n-1)}\bigl(D-R(v)-I(v)\bigr),
\label{eq:suppl_ce_fast_score}
\end{align}
with the same conventions for $n=0$ and $n=1$.
The implementation uses FP64 and omits the common factor $1/N$ in the coefficients during construction; this does not change candidate comparisons.

Algorithm~\ref{alg:suppl_ce} summarizes the construction.

\newcounter{supplalgorithm}
\noindent\begin{minipage}{\columnwidth}
\small
\newcommand{\loopline}[1]{\hspace*{0.75em}%
    \parbox[t]{\dimexpr\linewidth-0.75em\relax}{#1}}
\hrule
\vspace{3pt}
\refstepcounter{supplalgorithm}\label{alg:suppl_ce}
\noindent\textbf{Algorithm \thesupplalgorithm.} Global CE Order Construction
\par\vspace{3pt}
\hrule
\vspace{3pt}
\noindent\textbf{Input:} Influence matrix $W\in\mathbb{R}^{N\times N}$, window width $K$.\par
\noindent\textbf{Output:} Generation order $P$.\par
\begin{enumerate}[label=\arabic*:,leftmargin=1.65em,itemsep=1pt]
    \item Set $W_{a,a}\gets 0$ for every position $a$.
    \item Build $C$ using Eq.~\eqref{eq:suppl_ce_coeff}; precompute $H$ and $T$.
    \item $P\gets ()$, $U\gets\{1,\ldots,N\}$, $B\gets 0$.
    \item Initialize $R$, $I$, and $D$ from $W$ and $U$.
    \item \textbf{for} $m=1,\ldots,N$ \textbf{do}
    \item \loopline{Compute $Q_m(v)$ for all $v\in U$ using Eq.~\eqref{eq:suppl_ce_fast_score}, with $n=|U|-1$.}
    \item \loopline{$v^\star\gets\min\arg\min_{v\in U} Q_m(v)$.}
    \item \loopline{$B\gets B+\Delta_m(v^\star)$.}
    \item \loopline{$D\gets D-R(v^\star)-I(v^\star)$.}
    \item \loopline{$P\gets P\oplus(v^\star)$; $U\gets U\setminus\{v^\star\}$.}
    \item \loopline{For each $a$, set $R(a)\gets R(a)-W_{a,v^\star}$ and $I(a)\gets I(a)-W_{v^\star,a}$.}
    \item \textbf{end for}
    \item \textbf{return} $P$.
\end{enumerate}
\vspace{2pt}
\hrule
\end{minipage}
\par\medskip

\noindent\textbf{Complexity.}
The window loops construct $C$ in $O(NK^2)$ time; storing and initializing the dense coefficient matrix costs $O(N^2)$ space and time.
The implementation builds the row suffix sums in $O(N^2)$ time and computes $T_m$ by summing each dense suffix submatrix, taking $O(N^3)$ time in total.
At step $m$, the dense products used for candidate scoring take $O(m(N-m+1))$ time; updating the maintained sums takes $O(N)$.
The overall worst-case time is $O(N^3)$, and space is $O(N^2)$.

\noindent\textbf{Guarantee.}
Let $G(P)=\mathbb{E}[\mathcal{F}_K\mid P]$ for a uniformly completed prefix $P$.
For the nonempty remaining set $U$, the law of total expectation gives
\[
G(P)=\frac{1}{|U|}\sum_{v\in U}G(P\oplus v)
     \ge \min_{v\in U}G(P\oplus v).
\]
With exact conditional expectations, choosing the minimizing candidate makes $G(P)$ nonincreasing along the constructed prefix.
At the empty prefix, $G$ equals the uniform random-order expectation; at the complete prefix, it equals $\mathcal{F}_K(\pi_{\mathrm{SAO}})$.
This proves Eq.~(22) of the main paper.
The bound concerns the surrogate objective and does not establish optimality over permutations or any bound on the number of decoding rounds or latency.

\begin{figure}[t]
      \centering
      \includegraphics[width=\columnwidth]{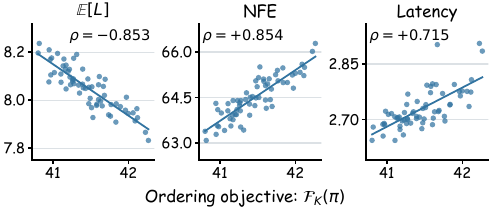}
      \caption{Ordering objective versus measured decoding efficiency
      on RLlamaGen at $K=16$.
      Each point represents one of 64 uniformly sampled generation orders.
      Lines show linear fits, and $\rho$ denotes Spearman's rank correlation.}
      \label{fig:ordering_objective_rllamagen}
\end{figure}

\section{Approximation Design and Justification}
\label{sec:suppl_approximations}

The original goal is to minimize the expected number of decoding rounds, $\mathbb{E}[T_\pi]$.
A round starts immediately after the tokens committed in the previous round, so the window starts visited during decoding depend on the generated history and earlier rejection outcomes.
Directly estimating $\mathbb{E}[T_\pi]$ for each candidate order would require repeated full decoding runs.
SAO makes four approximations to score candidate orders without those runs:

\begin{itemize}
    \item \textbf{Average over profiling histories.}
    For an eligible history $h$, we approximate its expected pairwise conditional influence using the global matrix:
    \begin{equation}
    \begin{aligned}
    &\mathbb{E}_{a\sim p_\theta(x_u\mid h)}\Big[
    \operatorname{TV}\big(
    p_\theta(x_v\mid h),\\
    &\hspace{4em}p_\theta(x_v\mid h\oplus(u,a))\big)\Big]
    \approx W_{u\rightarrow v}.
    \end{aligned}
    \label{eq:suppl_approx_history}
    \end{equation}
    As defined in the main paper, $W_{u\rightarrow v}$ averages this quantity over $h\sim\mathcal{D}^{uv}$, where neither position has been generated.
    Reusing this average avoids querying the model to estimate pairwise influence at each new history.

    \item \textbf{Add pairwise influences.}
    For a window beginning after $t$ confirmed tokens, we use the pairwise additive score
    \begin{equation}
    \bar{\delta}_j
    \approx \widehat{\delta}_{t,j}(\pi)
    =\sum_{i=1}^{j-1}W_{\pi_{t+i}\rightarrow\pi_{t+j}}.
    \label{eq:suppl_approx_pairwise}
    \end{equation}
    Here, $\bar{\delta}_j$ is the expected conditional mismatch given survival of the preceding drafts, as defined in the main paper, with the current history $c_t$ and order $\pi$ implicit.
    The score ignores higher-order interactions, cancellation between pairwise effects, and changes in the preceding-draft distribution induced by conditioning on acceptance.
    It avoids evaluating target distributions for combinations of preceding draft values and is neither an exact rejection probability nor a guaranteed upper bound.

    \item \textbf{Omit prefix survival.}
    With $k_t=\min(K,N-t)$, we approximate the expected progress loss by dropping the survival factors:
    \begin{equation}
    \begin{aligned}
    &\sum_{j=1}^{k_t-1}(k_t-j)S_{j-1}\bar{\delta}_j\\
    &\qquad\approx\sum_{j=1}^{k_t-1}(k_t-j)\bar{\delta}_j.
    \end{aligned}
    \label{eq:suppl_approx_survival}
    \end{equation}
    Here, $S_{j-1}=\Pr(A_{<j})$ is the probability that all preceding drafts are accepted.
    Substituting $\widehat{\delta}_{t,j}(\pi)$ for $\bar{\delta}_j$ gives the window-level surrogate $\mathcal{J}_K(\pi,t)$.
    This retains the larger cost of early rejection without estimating prefix survival for each candidate order.

    \item \textbf{Weight window starts uniformly.}
    Rather than estimating order-dependent visitation probabilities, we set
    \begin{equation}
    \begin{aligned}
    \rho_t&=\frac{1}{N},\qquad t=0,\ldots,N-1,\\
    \mathcal{F}_K(\pi)&=\frac{1}{N}\sum_{t=0}^{N-1}\mathcal{J}_K(\pi,t).
    \end{aligned}
    \label{eq:suppl_approx_window_weights}
    \end{equation}
    Uniform aggregation includes every possible window start without requiring full decoding runs to estimate how frequently each start is visited.
    It defines the global surrogate rather than an exact expression for the expected number of decoding rounds.
\end{itemize}

Together, these approximations make candidate scoring a calculation on the fixed matrix $W$, without running the generator or sampling completions at ordering time.
Fig.~3 of the main paper and \cref{fig:ordering_objective_rllamagen} show that the surrogate retains useful information for ranking generation orders by decoding efficiency on both backbones.

\begin{table*}[t]
    \centering
    \caption{ARPG-L results across speculative widths.
    Latency is measured in seconds per image with batch size one.
    Sequential Random shares quality and rollout metrics across widths;
    its latency is measured separately in each paired benchmark.
    NFE and latency speedups are relative to sequential decoding under the same order.
    Order speedup compares Self-Spec Random with Self-Spec SAO.
    Random-order results are reported as mean and standard deviation over five independently sampled generation orders.
    The best results among sequential and exact self-speculative methods are shown in \textbf{bold};
    Direct Parallel serves as a quality--efficiency reference.
    \textcolor{red}{Red} highlights the speedups of Self-Spec SAO.}
    \label{tab:suppl_k_scaling_arpg}
    \setlength{\tabcolsep}{4pt}
    \resizebox{\textwidth}{!}{%
    \begin{tabular}{cl*{7}{c}}
        \toprule
        \textbf{$K$} & \textbf{Method} & \textbf{NFE} $\downarrow$ &
        $\boldsymbol{\mathbb{E}[L]}$ $\uparrow$ & \textbf{Latency (s)} $\downarrow$ &
        \multicolumn{3}{c}{\textbf{Speedup} $\uparrow$} & \textbf{FID} $\downarrow$ \\
        \cmidrule(lr){6-8}
        & & & & & \textbf{NFE} & \textbf{Latency} & \textbf{Order} & \\
        \midrule
        $4$ & Sequential Random & 256.00 & 1.00 & $4.80{\scriptstyle\,\pm\,0.01}$ & $1.00\times$ & $1.00\times$ & -- & $16.89{\scriptstyle\,\pm\,0.14}$ \\
         & Sequential SAO & 256.00 & 1.00 & 4.79 & $1.00\times$ & $1.00\times$ & -- & \textbf{15.05} \\
        \addlinespace[5pt]
         & Self-Spec Random & $132.57{\scriptstyle\,\pm\,0.01}$ & $3.85{\scriptstyle\,\pm\,0.00}$ & $2.57{\scriptstyle\,\pm\,0.00}$ & $1.93{\scriptstyle\,\pm\,0.00}\times$ & $1.86{\scriptstyle\,\pm\,0.00}\times$ & -- & $16.89{\scriptstyle\,\pm\,0.06}$ \\
         & \textbf{Self-Spec SAO} & \textbf{130.07} & \textbf{3.93} & \textbf{2.52} & \textcolor{red}{$\mathbf{1.97}\times$} & \textcolor{red}{$\mathbf{1.90}\times$} & \textcolor{red}{$\mathbf{1.02}\times$} & 15.06 \\
        \cmidrule(lr){2-9}
         & Direct Parallel & 64.00 & 4.00 & $1.27{\scriptstyle\,\pm\,0.00}$ & $4.00\times$ & $3.78{\scriptstyle\,\pm\,0.00}\times$ & -- & $20.89{\scriptstyle\,\pm\,0.03}$ \\
        \midrule
        $8$ & Sequential Random & 256.00 & 1.00 & $4.79{\scriptstyle\,\pm\,0.00}$ & $1.00\times$ & $1.00\times$ & -- & $16.89{\scriptstyle\,\pm\,0.14}$ \\
         & Sequential SAO & 256.00 & 1.00 & 4.81 & $1.00\times$ & $1.00\times$ & -- & \textbf{14.41} \\
        \addlinespace[5pt]
         & Self-Spec Random & $76.45{\scriptstyle\,\pm\,0.01}$ & $6.68{\scriptstyle\,\pm\,0.00}$ & $1.50{\scriptstyle\,\pm\,0.01}$ & $3.35{\scriptstyle\,\pm\,0.00}\times$ & $3.20{\scriptstyle\,\pm\,0.02}\times$ & -- & $17.01{\scriptstyle\,\pm\,0.04}$ \\
         & \textbf{Self-Spec SAO} & \textbf{70.23} & \textbf{7.28} & \textbf{1.39} & \textcolor{red}{$\mathbf{3.64}\times$} & \textcolor{red}{$\mathbf{3.46}\times$} & \textcolor{red}{$\mathbf{1.08}\times$} & 14.60 \\
        \cmidrule(lr){2-9}
         & Direct Parallel & 32.00 & 8.00 & $0.65{\scriptstyle\,\pm\,0.00}$ & $8.00\times$ & $7.39{\scriptstyle\,\pm\,0.01}\times$ & -- & $26.26{\scriptstyle\,\pm\,0.28}$ \\
        \midrule
        $16$ & Sequential Random & 256.00 & 1.00 & $4.82{\scriptstyle\,\pm\,0.01}$ & $1.00\times$ & $1.00\times$ & -- & $16.89{\scriptstyle\,\pm\,0.14}$ \\
         & Sequential SAO & 256.00 & 1.00 & 4.78 & $1.00\times$ & $1.00\times$ & -- & \textbf{14.28} \\
        \addlinespace[5pt]
         & Self-Spec Random & $56.46{\scriptstyle\,\pm\,0.02}$ & $9.05{\scriptstyle\,\pm\,0.00}$ & $1.12{\scriptstyle\,\pm\,0.03}$ & $4.53{\scriptstyle\,\pm\,0.00}\times$ & $4.30{\scriptstyle\,\pm\,0.11}\times$ & -- & $16.99{\scriptstyle\,\pm\,0.01}$ \\
         & \textbf{Self-Spec SAO} & \textbf{45.60} & \textbf{11.21} & \textbf{0.90} & \textcolor{red}{$\mathbf{5.61}\times$} & \textcolor{red}{$\mathbf{5.29}\times$} & \textcolor{red}{$\mathbf{1.24}\times$} & 14.67 \\
        \cmidrule(lr){2-9}
         & Direct Parallel & 16.00 & 16.00 & $0.33{\scriptstyle\,\pm\,0.00}$ & $16.00\times$ & $14.40{\scriptstyle\,\pm\,0.02}\times$ & -- & $38.28{\scriptstyle\,\pm\,0.19}$ \\
        \midrule
        $32$ & Sequential Random & 256.00 & 1.00 & $4.82{\scriptstyle\,\pm\,0.04}$ & $1.00\times$ & $1.00\times$ & -- & $16.89{\scriptstyle\,\pm\,0.14}$ \\
         & Sequential SAO & 256.00 & 1.00 & 4.79 & $1.00\times$ & $1.00\times$ & -- & 15.28 \\
        \addlinespace[5pt]
         & Self-Spec Random & $53.63{\scriptstyle\,\pm\,0.06}$ & $9.53{\scriptstyle\,\pm\,0.01}$ & $1.09{\scriptstyle\,\pm\,0.02}$ & $4.77{\scriptstyle\,\pm\,0.00}\times$ & $4.45{\scriptstyle\,\pm\,0.07}\times$ & -- & $16.97{\scriptstyle\,\pm\,0.09}$ \\
         & \textbf{Self-Spec SAO} & \textbf{40.50} & \textbf{12.62} & \textbf{0.81} & \textcolor{red}{$\mathbf{6.32}\times$} & \textcolor{red}{$\mathbf{5.92}\times$} & \textcolor{red}{$\mathbf{1.34}\times$} & \textbf{15.15} \\
        \cmidrule(lr){2-9}
         & Direct Parallel & 8.00 & 32.00 & $0.18{\scriptstyle\,\pm\,0.00}$ & $32.00\times$ & $27.28{\scriptstyle\,\pm\,0.02}\times$ & -- & $60.71{\scriptstyle\,\pm\,0.18}$ \\
        \midrule
        $64$ & Sequential Random & 256.00 & 1.00 & $4.85{\scriptstyle\,\pm\,0.02}$ & $1.00\times$ & $1.00\times$ & -- & $16.89{\scriptstyle\,\pm\,0.14}$ \\
         & Sequential SAO & 256.00 & 1.00 & 4.83 & $1.00\times$ & $1.00\times$ & -- & 15.29 \\
        \addlinespace[5pt]
         & Self-Spec Random & $53.62{\scriptstyle\,\pm\,0.05}$ & $9.53{\scriptstyle\,\pm\,0.01}$ & $1.08{\scriptstyle\,\pm\,0.02}$ & $4.77{\scriptstyle\,\pm\,0.00}\times$ & $4.48{\scriptstyle\,\pm\,0.09}\times$ & -- & $16.91{\scriptstyle\,\pm\,0.07}$ \\
         & \textbf{Self-Spec SAO} & \textbf{41.24} & \textbf{12.40} & \textbf{0.86} & \textcolor{red}{$\mathbf{6.21}\times$} & \textcolor{red}{$\mathbf{5.65}\times$} & \textcolor{red}{$\mathbf{1.26}\times$} & \textbf{15.17} \\
        \cmidrule(lr){2-9}
         & Direct Parallel & 4.00 & 64.00 & $0.10{\scriptstyle\,\pm\,0.00}$ & $64.00\times$ & $49.39{\scriptstyle\,\pm\,0.13}\times$ & -- & $94.96{\scriptstyle\,\pm\,0.23}$ \\
        \bottomrule
    \end{tabular}%
    }
\end{table*}

\begin{table*}[t]
    \centering
    \caption{RLlamaGen results across speculative widths.
    Latency is measured in seconds per image with batch size one.
    Sequential Random shares quality and rollout metrics across widths;
    its latency is measured separately in each paired benchmark.
    NFE and latency speedups are relative to sequential decoding under the same order.
    Order speedup compares Self-Spec Random with Self-Spec SAO.
    Random-order results are reported as mean and standard deviation over independently sampled generation orders.
    The best results among sequential and exact self-speculative methods are shown in \textbf{bold};
    Direct Parallel serves as a quality--efficiency reference.
    \textcolor{red}{Red} highlights the speedups of Self-Spec SAO.}
    \label{tab:suppl_k_scaling_rllamagen}
    \setlength{\tabcolsep}{4pt}
    \resizebox{\textwidth}{!}{%
    \begin{tabular}{cl*{8}{c}}
        \toprule
        \textbf{$K$} & \textbf{Method} & \textbf{NFE} $\downarrow$ &
        $\boldsymbol{\mathbb{E}[L]}$ $\uparrow$ & \textbf{Latency (s)} $\downarrow$ &
        \multicolumn{3}{c}{\textbf{Speedup} $\uparrow$} & \textbf{FID} $\downarrow$ & \textbf{CLIP-Score} $\uparrow$ \\
        \cmidrule(lr){6-8}
        & & & & & \textbf{NFE} & \textbf{Latency} & \textbf{Order} & & \\
        \midrule
        $4$ & Sequential Random & 257.00 & 1.00 & $10.58{\scriptstyle\,\pm\,0.11}$ & $1.00\times$ & $1.00\times$ & -- & $31.03{\scriptstyle\,\pm\,0.05}$ & $27.34{\scriptstyle\,\pm\,0.01}$ \\
         & Sequential SAO & 257.00 & 1.00 & 10.54 & $1.00\times$ & $1.00\times$ & -- & 30.34 & 27.37 \\
        \addlinespace[5pt]
         & Self-Spec Random & $137.88{\scriptstyle\,\pm\,0.02}$ & $3.73{\scriptstyle\,\pm\,0.00}$ & $5.73{\scriptstyle\,\pm\,0.07}$ & $1.86{\scriptstyle\,\pm\,0.00}\times$ & $1.85{\scriptstyle\,\pm\,0.00}\times$ & -- & $31.11{\scriptstyle\,\pm\,0.13}$ & $27.35{\scriptstyle\,\pm\,0.01}$ \\
         & \textbf{Self-Spec SAO} & \textbf{136.38} & \textbf{3.77} & \textbf{5.65} & \textcolor{red}{$\mathbf{1.88}\times$} & \textcolor{red}{$\mathbf{1.87}\times$} & \textcolor{red}{$\mathbf{1.02}\times$} & \textbf{29.91} & \textbf{27.44} \\
        \cmidrule(lr){2-10}
         & Direct Parallel & 129.00 & 4.00 & $5.32{\scriptstyle\,\pm\,0.06}$ & $1.99\times$ & $1.99{\scriptstyle\,\pm\,0.00}\times$ & -- & $33.25{\scriptstyle\,\pm\,0.15}$ & $27.14{\scriptstyle\,\pm\,0.03}$ \\
        \midrule
        $8$ & Sequential Random & 257.00 & 1.00 & $10.60{\scriptstyle\,\pm\,0.02}$ & $1.00\times$ & $1.00\times$ & -- & $31.03{\scriptstyle\,\pm\,0.05}$ & $27.34{\scriptstyle\,\pm\,0.01}$ \\
         & Sequential SAO & 257.00 & 1.00 & 10.56 & $1.00\times$ & $1.00\times$ & -- & \textbf{30.19} & \textbf{27.39} \\
        \addlinespace[5pt]
         & Self-Spec Random & $83.86{\scriptstyle\,\pm\,0.02}$ & $6.17{\scriptstyle\,\pm\,0.00}$ & $3.54{\scriptstyle\,\pm\,0.03}$ & $3.06{\scriptstyle\,\pm\,0.00}\times$ & $2.99{\scriptstyle\,\pm\,0.02}\times$ & -- & $30.98{\scriptstyle\,\pm\,0.08}$ & $27.34{\scriptstyle\,\pm\,0.02}$ \\
         & \textbf{Self-Spec SAO} & \textbf{79.33} & \textbf{6.53} & \textbf{3.34} & \textcolor{red}{$\mathbf{3.24}\times$} & \textcolor{red}{$\mathbf{3.17}\times$} & \textcolor{red}{$\mathbf{1.06}\times$} & \textbf{30.19} & \textbf{27.39} \\
        \cmidrule(lr){2-10}
         & Direct Parallel & 65.00 & 8.00 & $2.75{\scriptstyle\,\pm\,0.01}$ & $3.95\times$ & $3.86{\scriptstyle\,\pm\,0.01}\times$ & -- & $36.69{\scriptstyle\,\pm\,0.09}$ & $26.90{\scriptstyle\,\pm\,0.02}$ \\
        \midrule
        $16$ & Sequential Random & 257.00 & 1.00 & $10.52{\scriptstyle\,\pm\,0.01}$ & $1.00\times$ & $1.00\times$ & -- & $31.03{\scriptstyle\,\pm\,0.05}$ & $27.34{\scriptstyle\,\pm\,0.01}$ \\
         & Sequential SAO & 257.00 & 1.00 & 10.21 & $1.00\times$ & $1.00\times$ & -- & 30.40 & 27.39 \\
        \addlinespace[5pt]
         & Self-Spec Random & $64.34{\scriptstyle\,\pm\,0.09}$ & $8.07{\scriptstyle\,\pm\,0.01}$ & $2.73{\scriptstyle\,\pm\,0.03}$ & $3.99{\scriptstyle\,\pm\,0.01}\times$ & $3.85{\scriptstyle\,\pm\,0.04}\times$ & -- & $30.93{\scriptstyle\,\pm\,0.21}$ & $27.34{\scriptstyle\,\pm\,0.00}$ \\
         & \textbf{Self-Spec SAO} & \textbf{55.54} & \textbf{9.37} & \textbf{2.25} & \textcolor{red}{$\mathbf{4.63}\times$} & \textcolor{red}{$\mathbf{4.53}\times$} & \textcolor{red}{$\mathbf{1.21}\times$} & \textbf{30.33} & \textbf{27.42} \\
        \cmidrule(lr){2-10}
         & Direct Parallel & 33.00 & 16.00 & $1.39{\scriptstyle\,\pm\,0.00}$ & $7.79\times$ & $7.55{\scriptstyle\,\pm\,0.02}\times$ & -- & $43.95{\scriptstyle\,\pm\,0.16}$ & $26.33{\scriptstyle\,\pm\,0.03}$ \\
        \midrule
        $32$ & Sequential Random & 257.00 & 1.00 & $10.53{\scriptstyle\,\pm\,0.06}$ & $1.00\times$ & $1.00\times$ & -- & $31.03{\scriptstyle\,\pm\,0.05}$ & $27.34{\scriptstyle\,\pm\,0.01}$ \\
         & Sequential SAO & 257.00 & 1.00 & 10.52 & $1.00\times$ & $1.00\times$ & -- & \textbf{30.03} & \textbf{27.42} \\
        \addlinespace[5pt]
         & Self-Spec Random & $61.52{\scriptstyle\,\pm\,0.06}$ & $8.45{\scriptstyle\,\pm\,0.01}$ & $2.60{\scriptstyle\,\pm\,0.01}$ & $4.18{\scriptstyle\,\pm\,0.00}\times$ & $4.05{\scriptstyle\,\pm\,0.00}\times$ & -- & $30.96{\scriptstyle\,\pm\,0.19}$ & $27.35{\scriptstyle\,\pm\,0.01}$ \\
         & \textbf{Self-Spec SAO} & \textbf{50.45} & \textbf{10.34} & \textbf{2.07} & \textcolor{red}{$\mathbf{5.09}\times$} & \textcolor{red}{$\mathbf{5.09}\times$} & \textcolor{red}{$\mathbf{1.26}\times$} & 30.40 & 27.39 \\
        \cmidrule(lr){2-10}
         & Direct Parallel & 17.00 & 32.00 & $0.73{\scriptstyle\,\pm\,0.01}$ & $15.12\times$ & $14.50{\scriptstyle\,\pm\,0.02}\times$ & -- & $59.35{\scriptstyle\,\pm\,0.39}$ & $25.15{\scriptstyle\,\pm\,0.03}$ \\
        \midrule
        $64$ & Sequential Random & 257.00 & 1.00 & $10.55{\scriptstyle\,\pm\,0.01}$ & $1.00\times$ & $1.00\times$ & -- & $31.03{\scriptstyle\,\pm\,0.05}$ & $27.34{\scriptstyle\,\pm\,0.01}$ \\
         & Sequential SAO & 257.00 & 1.00 & 10.52 & $1.00\times$ & $1.00\times$ & -- & \textbf{30.49} & 27.37 \\
        \addlinespace[5pt]
         & Self-Spec Random & $61.54{\scriptstyle\,\pm\,0.03}$ & $8.44{\scriptstyle\,\pm\,0.00}$ & $2.64{\scriptstyle\,\pm\,0.05}$ & $4.18{\scriptstyle\,\pm\,0.00}\times$ & $4.00{\scriptstyle\,\pm\,0.09}\times$ & -- & $30.87{\scriptstyle\,\pm\,0.03}$ & $27.35{\scriptstyle\,\pm\,0.01}$ \\
         & \textbf{Self-Spec SAO} & \textbf{51.46} & \textbf{10.13} & \textbf{2.15} & \textcolor{red}{$\mathbf{4.99}\times$} & \textcolor{red}{$\mathbf{4.89}\times$} & \textcolor{red}{$\mathbf{1.23}\times$} & 30.67 & \textbf{27.39} \\
        \cmidrule(lr){2-10}
         & Direct Parallel & 9.00 & 64.00 & $0.40{\scriptstyle\,\pm\,0.00}$ & $28.56\times$ & $26.64{\scriptstyle\,\pm\,0.14}\times$ & -- & $87.81{\scriptstyle\,\pm\,0.08}$ & $23.46{\scriptstyle\,\pm\,0.02}$ \\
        \bottomrule
    \end{tabular}%
    }
\end{table*}

\begin{table*}[t]
    \centering
    \caption{Effect of profiling budget on ARPG-L and RLlamaGen at $K=16$.
    Coverage denotes the fraction of observed off-diagonal position pairs.
    Profiling cost includes history generation, conditional-TV estimation, and matrix aggregation, measured in GPU hours on an A100 40GB.
    Random serves as the baseline without profiling.
    Dashes at the 64-trajectory budget indicate that order construction and rollout are omitted due to incomplete matrix coverage.}
    \label{tab:profiling_budget}
    \setlength{\tabcolsep}{5pt}
    \begin{tabular}{clcccc}
        \toprule
        \textbf{Backbone} & \textbf{Profiling Trajectories} & \textbf{Coverage (\%)} $\uparrow$ & $\mathbb{E}[L]\uparrow$ & \textbf{NFE} $\downarrow$ & \textbf{Profiling Cost (GPU hours)} $\downarrow$ \\
        \midrule
        \multirow{8}{*}{ARPG-L} & Random & -- & $9.05{\scriptstyle\,\pm\,0.00}$ & $56.46{\scriptstyle\,\pm\,0.02}$ & 0.00 \\
        \addlinespace[2pt]
         & 64 & 99.98 & -- & -- & 0.12 \\
         & 128 & 100.00 & 11.06 & 46.23 & 0.24 \\
         & 256 & 100.00 & 11.07 & 46.15 & 0.47 \\
         & 500 & 100.00 & 11.16 & 45.82 & 0.92 \\
         & 1,000 & 100.00 & 11.16 & 45.79 & 1.84 \\
         & 2,000 & 100.00 & 11.20 & 45.66 & 3.67 \\
         & 5,000 & 100.00 & \textbf{11.21} & \textbf{45.60} & 9.18 \\
        \midrule
        \multirow{8}{*}{RLlamaGen} & Random & -- & $8.07{\scriptstyle\,\pm\,0.01}$ & $64.34{\scriptstyle\,\pm\,0.09}$ & 0.00 \\
        \addlinespace[2pt]
         & 64 & 99.95 & -- & -- & 0.20 \\
         & 128 & 100.00 & 9.28 & 56.02 & 0.41 \\
         & 256 & 100.00 & 9.35 & 55.66 & 0.82 \\
         & 500 & 100.00 & 9.32 & 55.84 & 1.59 \\
         & 1,000 & 100.00 & 9.34 & 55.69 & 3.16 \\
         & 2,000 & 100.00 & 9.33 & 55.75 & 6.30 \\
         & 5,000 & 100.00 & \textbf{9.37} & \textbf{55.54} & 13.95 \\
        \bottomrule
    \end{tabular}%
\end{table*}

\section{Detailed Experimental Setup}
\label{sec:suppl_setup}

\noindent\textbf{Offline Matrix Construction.}
Profiling histories are sampled from complete serial trajectories generated by the frozen model under random orders.
For each trajectory, we sample one prefix length uniformly from each of $[0,63]$, $[64,127]$, $[128,191]$, and $[192,254]$.
At each history, we select up to 16 anchors uniformly from the remaining positions and sample one value per anchor from its model distribution.
Each anchor is temporarily appended to the original history to measure the conditional TV at the other remaining positions.
We combine observations across the four stages using inverse anchor-inclusion weights, $|U|/\min(16,|U|)$, where $U$ is the remaining position set.
The resulting global matrix is directed; we use no spatial prior, smoothing, or missing-value imputation.

\noindent\textbf{Decoding and Numerical Settings.}
We use the with-bonus protocol: a window of actual width $k_t=\min(K,N-t)$ contains up to $k_t-1$ draft proposals and one potential bonus token.
The complete generation order is constructed offline and remains fixed throughout decoding.
After rejection, only the accepted prefix and a residual correction token are committed, and decoding continues along the same order.
The cache retains only confirmed history, discarding uncommitted draft states.
Model forwards use FP32 with math SDPA and TF32 disabled; probability sampling, TV estimation, and order construction use FP64.

\noindent\textbf{Metrics and Timing.}
For $B$ images, let $T_b$ be the number of decoding rounds, $L_{b,r}$ the committed length of round $r$, and $n_b$ the logical model-evaluation count for image $b$.
We aggregate committed length over rounds and NFE over images:
\begin{equation}
\mathbb{E}[L]
=\frac{\sum_{b=1}^{B}\sum_{r=1}^{T_b}L_{b,r}}
       {\sum_{b=1}^{B}T_b},
\qquad
\mathrm{NFE}=\frac{1}{B}\sum_{b=1}^{B}n_b.
\label{eq:suppl_efficiency_metrics}
\end{equation}
For a decoder $D$ and a fixed order $\pi$, the sequential-reference speedups are
\begin{equation}
\begin{aligned}
s_{\mathrm{NFE}}(D,\pi)
&=\frac{\mathrm{NFE}(\mathrm{Seq},\pi)}
        {\mathrm{NFE}(D,\pi)},\\
s_{\mathrm{lat}}(D,\pi)
&=\frac{\tau(\mathrm{Seq},\pi)}
        {\tau(D,\pi)},
\end{aligned}
\label{eq:suppl_sequential_speedups}
\end{equation}
where $\tau$ denotes mean latency per image.
The latency benefit of an alternative generation order $\pi$ is
\begin{equation}
s_{\mathrm{order}}(D,\pi)
=\frac{\tau(D,\mathrm{Random})}
       {\tau(D,\pi)}.
\label{eq:suppl_order_speedup}
\end{equation}
Here, $\tau(D,\mathrm{Random})$ is the mean latency over random-order runs, and $\tau(D,\pi)$ is the latency of the same decoder under the compared order $\pi$.
NFE counts parallel queries as one logical evaluation.
RLlamaGen additionally counts text-prefix initialization, and its Direct Parallel decoder counts the causal cache-materialization pass.
Latency benchmarks use batch size one on an exclusive A100 40GB, with 32 warmup samples followed by 64 timed samples and GPU synchronization at timing boundaries.
The timed scope includes prefix-cache initialization, decoding and cache operations, and VQ reconstruction, excluding model loading, text-feature extraction, and disk I/O.

\begin{figure*}[t]
      \centering
      \includegraphics[width=\textwidth]{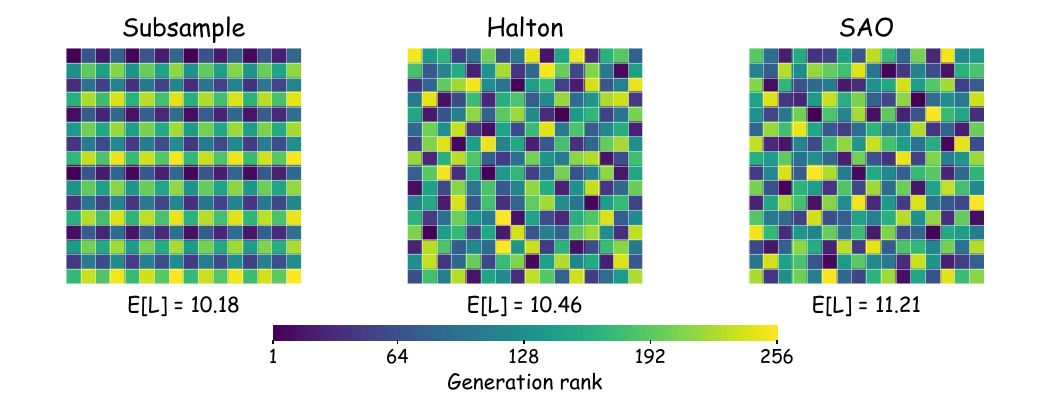}
      \caption{Visualization of Subsample, Halton, and SAO generation
      orders on the $16\times16$ token grid of ARPG-L at $K=16$.
      Colors indicate generation ranks, with darker positions generated
      earlier and brighter positions generated later.}
      \label{fig:generation_orders}
\end{figure*}

\section{Results Across Speculative Widths}
\label{sec:suppl_k_scaling}

Tables~\ref{tab:suppl_k_scaling_arpg} and~\ref{tab:suppl_k_scaling_rllamagen} provide the complete numerical results across speculative widths, including sequential and Direct Parallel baselines.
Across these settings, Self-Spec SAO offers a consistent balance between generation quality and decoding efficiency.

\section{Offline Budget}
\label{sec:suppl_profiling_budget}

We examine how the number of profiling trajectories affects matrix coverage, decoding efficiency, and profiling cost on both backbones at $K=16$ and CFG=1.
All budgets follow the same sampling protocol: each trajectory contributes four history prefixes, with one prefix length sampled uniformly from each of $[0,63]$, $[64,127]$, $[128,191]$, and $[192,254]$.
For each prefix, we uniformly select up to 16 anchor positions from the remaining positions and sample one token value per anchor from its model distribution.
We fix the trajectory ordering and use the first 64, 128, 256, 500, 1,000, 2,000, or 5,000 trajectories to construct each matrix.
Only the trajectory budget changes; matrix estimation, order construction, and decoding settings remain fixed.

As shown in \cref{tab:profiling_budget}, 64 trajectories do not cover all off-diagonal position pairs, so we omit order construction and rollout at this budget.
Both backbones achieve full coverage with 128 trajectories.
Relatively small budgets already produce effective orders: with 500 trajectories, ARPG-L achieves an average committed length of 11.16 and an NFE of 45.82, close to the main-result 5K configuration (11.21 and 45.60).
With 128 trajectories, RLlamaGen achieves 9.28 and 56.02, respectively, compared with 9.37 and 55.54 at 5K.
Further increases in budget yield smaller changes in decoding efficiency, although full coverage alone does not establish statistical convergence of the matrix estimates.

The reported profiling cost includes history generation, conditional-TV estimation, and matrix aggregation, measured in GPU hours.
The 500-trajectory configuration on ARPG-L costs 0.92 GPU hours, while the 128-trajectory configuration on RLlamaGen costs 0.41 GPU hours.
Thus, in these experiments, less than one GPU hour of profiling produces decoding efficiency close to that of the 5K configuration.

We construct the complete 256-token generation order on the CPU using the method of conditional expectations.
At $K=16$, construction from the 5K-trajectory matrices takes 47.54\,ms for ARPG-L and 47.31\,ms for RLlamaGen.
These CPU construction times are separate from the profiling costs reported in the table.
The resulting generation order can be reused, requiring no additional model evaluations or online order selection during generation.

\begin{table*}[t]
    \centering
     \caption{Generation-order comparison on ARPG-L at $K=16$. FID is evaluated on 50K ImageNet images. Latency is reported in seconds per image at batch size one. Order speedup is the ratio of Random latency to the latency of the listed order. Stochastic orders are evaluated with five independent order seeds. The best result for each metric is shown in \textbf{bold}.}
    \label{tab:order_comparison_arpg}
    \setlength{\tabcolsep}{3pt}
    
    \begin{tabular}{l*{5}{c}}
        \toprule
        \textbf{Generation Order} & $\mathbb{E}[L]\uparrow$ & NFE $\downarrow$ & Latency $\downarrow$ & Order Speedup $\uparrow$ & FID $\downarrow$ \\
        \midrule
        Random & $9.05{\scriptstyle\,\pm\,0.00}$ & $56.46{\scriptstyle\,\pm\,0.02}$ & $1.12{\scriptstyle\,\pm\,0.03}$ & $1.00\times$ & $16.99{\scriptstyle\,\pm\,0.01}$ \\
        \addlinespace[4pt]
        Raster Scan~\cite{esser2021taming} & 4.14 & 123.32 & 2.39 & $0.47\times$ & 41.00 \\
        Spiral-in~\cite{esser2021taming} & 4.07 & 125.66 & 2.34 & $0.48\times$ & 35.50 \\
        Spiral-out~\cite{esser2021taming} & 3.91 & 130.57 & 2.46 & $0.46\times$ & 34.81 \\
        Z-curve~\cite{esser2021taming} & 4.20 & 121.62 & 2.32 & $0.48\times$ & 43.42 \\
        Subsample~\cite{esser2021taming} & 10.18 & 50.20 & 1.01 & $1.11\times$ & 18.93 \\
        Alternate~\cite{esser2021taming} & 3.95 & 129.48 & 2.49 & $0.45\times$ & 40.76 \\
        Halton~\cite{halton1960efficiency} & 10.46 & 48.88 & 0.97 & $1.15\times$ & 15.02 \\
        \addlinespace[4pt]
        Hierarchical Random~\cite{pang2025randar} & $9.11{\scriptstyle\,\pm\,0.09}$ & $56.14{\scriptstyle\,\pm\,0.55}$ & $1.07{\scriptstyle\,\pm\,0.02}$ & $1.05{\scriptstyle\,\pm\,0.01}\times$ & $15.86{\scriptstyle\,\pm\,0.30}$ \\
        Center-first Random~\cite{pang2025randar} & $8.24{\scriptstyle\,\pm\,0.13}$ & $62.04{\scriptstyle\,\pm\,0.85}$ & $1.20{\scriptstyle\,\pm\,0.02}$ & $0.93{\scriptstyle\,\pm\,0.01}\times$ & $23.87{\scriptstyle\,\pm\,0.73}$ \\
        Border-first Random~\cite{pang2025randar} & $8.13{\scriptstyle\,\pm\,0.07}$ & $62.85{\scriptstyle\,\pm\,0.54}$ & $1.23{\scriptstyle\,\pm\,0.03}$ & $0.91{\scriptstyle\,\pm\,0.01}\times$ & $19.47{\scriptstyle\,\pm\,1.16}$ \\
        \addlinespace[4pt]
        PAR-inspired~\cite{wang2025par} & 6.83 & 74.78 & 1.44 & $0.78\times$ & 27.63 \\
        LPD-inspired~\cite{zhang2026locality} & $9.12{\scriptstyle\,\pm\,0.13}$ & $56.01{\scriptstyle\,\pm\,0.86}$ & $1.09{\scriptstyle\,\pm\,0.02}$ & $1.03{\scriptstyle\,\pm\,0.01}\times$ & $17.83{\scriptstyle\,\pm\,0.63}$ \\
        \addlinespace[4pt]
        \textbf{SAO (Ours)} & \textbf{11.21} & \textbf{45.60} & \textbf{0.90} & $\mathbf{1.24\times}$ & \textbf{14.67} \\
        \bottomrule
    \end{tabular}%
\end{table*}

\begin{table*}[t]
    \centering
    \caption{Generation-order comparison on RLlamaGen at $K=16$. FID is evaluated on 30K COCO images. 
    CLIP denotes CLIP cosine similarity multiplied by 100.
    Latency is reported in seconds per image at batch size one.
    Order speedup is the ratio of Random latency to the latency of the listed order.
    Stochastic orders are evaluated with five independent order seeds.
    The best result for each metric is shown in \textbf{bold}.}

    \label{tab:order_comparison_rllamagen}
    \setlength{\tabcolsep}{3pt}
    
    \begin{tabular}{l*{6}{c}}
        \toprule
        \textbf{Generation Order} & $\mathbb{E}[L]\uparrow$ & NFE $\downarrow$ & Latency $\downarrow$ & Order Speedup $\uparrow$ & FID $\downarrow$ & CLIP $\uparrow$ \\
        \midrule
        Random & $8.07{\scriptstyle\,\pm\,0.01}$ & $64.34{\scriptstyle\,\pm\,0.09}$ & $2.73{\scriptstyle\,\pm\,0.03}$ & $1.00\times$ & $30.93{\scriptstyle\,\pm\,0.21}$ & $27.34{\scriptstyle\,\pm\,0.00}$ \\
        \addlinespace[4pt]
        Raster Scan~\cite{esser2021taming} & 4.00 & 128.60 & 5.43 & $0.50\times$ & 34.57 & 26.49 \\
        Spiral-in~\cite{esser2021taming} & 4.04 & 127.54 & 5.45 & $0.50\times$ & 33.96 & 26.61 \\
        Spiral-out~\cite{esser2021taming} & 4.02 & 128.24 & 5.46 & $0.50\times$ & 35.31 & 26.66 \\
        Z-curve~\cite{esser2021taming} & 4.08 & 126.15 & 5.33 & $0.51\times$ & 33.52 & 26.64 \\
        Subsample~\cite{esser2021taming} & 8.73 & 59.55 & 2.54 & $1.07\times$ & 30.89 & 27.29 \\
        Alternate~\cite{esser2021taming} & 3.88 & 132.88 & 5.63 & $0.48\times$ & 34.82 & 26.51 \\
        Halton~\cite{halton1960efficiency} & 8.79 & 59.26 & 2.54 & $1.07\times$ & 30.60 & 27.19 \\
        \addlinespace[4pt]
        Hierarchical Random~\cite{pang2025randar} & $8.14{\scriptstyle\,\pm\,0.07}$ & $63.81{\scriptstyle\,\pm\,0.59}$ & $2.64{\scriptstyle\,\pm\,0.06}$ & $1.04{\scriptstyle\,\pm\,0.02}\times$ & $30.72{\scriptstyle\,\pm\,0.20}$ & $27.36{\scriptstyle\,\pm\,0.04}$ \\
        Center-first Random~\cite{pang2025randar} & $7.54{\scriptstyle\,\pm\,0.08}$ & $68.76{\scriptstyle\,\pm\,0.61}$ & $2.83{\scriptstyle\,\pm\,0.07}$ & $0.96{\scriptstyle\,\pm\,0.02}\times$ & $33.17{\scriptstyle\,\pm\,0.18}$ & $27.05{\scriptstyle\,\pm\,0.03}$ \\
        Border-first Random~\cite{pang2025randar} & $7.46{\scriptstyle\,\pm\,0.07}$ & $69.45{\scriptstyle\,\pm\,0.62}$ & $2.88{\scriptstyle\,\pm\,0.03}$ & $0.95{\scriptstyle\,\pm\,0.01}\times$ & $32.29{\scriptstyle\,\pm\,0.20}$ & $27.15{\scriptstyle\,\pm\,0.05}$ \\
        \addlinespace[4pt]
        PAR-inspired~\cite{wang2025par} & 6.15 & 84.11 & 3.58 & $0.76\times$ & 35.10 & 26.76 \\
        LPD-inspired~\cite{zhang2026locality} & $8.12{\scriptstyle\,\pm\,0.08}$ & $63.91{\scriptstyle\,\pm\,0.64}$ & $2.66{\scriptstyle\,\pm\,0.09}$ & $1.03{\scriptstyle\,\pm\,0.03}\times$ & $32.44{\scriptstyle\,\pm\,0.15}$ & $27.15{\scriptstyle\,\pm\,0.04}$ \\
        \addlinespace[4pt]
        \textbf{SAO (Ours)} & \textbf{9.37} & \textbf{55.54} & \textbf{2.25} & $\mathbf{1.21\times}$ & \textbf{30.33} & \textbf{27.42} \\
        \bottomrule
    \end{tabular}%
\end{table*}

\begin{table*}[t]
    \centering
    \caption{Effect of classifier-free guidance on both backbones at $K=16$.
    SAO (Reuse) applies the order constructed at CFG=1, while SAO (CFG-matched) reconstructs the order using 5K profiling trajectories at the corresponding guidance scale.
    Random reports means and standard deviations over five order seeds.
    The best result within each backbone and guidance scale is shown in \textbf{bold}.}
    \label{tab:cfg_comparison}
    \setlength{\tabcolsep}{3pt}
    \begin{tabular}{cclcccc}
        \toprule
        \textbf{Backbone} & \textbf{CFG} & \textbf{Order} & $\mathbb{E}[L]\uparrow$ & \textbf{NFE} $\downarrow$ & \textbf{FID} $\downarrow$ & \textbf{CLIP} $\uparrow$ \\
        \midrule
        \multirow{9}{*}{ARPG-L} & \multirow{3}{*}{2} & Random & $8.51{\scriptstyle\,\pm\,0.02}$ & $60.09{\scriptstyle\,\pm\,0.15}$ & $5.11{\scriptstyle\,\pm\,0.20}$ & -- \\
         &  & SAO (Reuse) & \textbf{10.54} & \textbf{48.51} & 5.80 & -- \\
         &  & SAO (CFG-matched) & 10.53 & 48.54 & \textbf{5.06} & -- \\
        \cmidrule(lr){2-7}
         & \multirow{3}{*}{3} & Random & $8.02{\scriptstyle\,\pm\,0.04}$ & $63.72{\scriptstyle\,\pm\,0.29}$ & $10.61{\scriptstyle\,\pm\,0.28}$ & -- \\
         &  & SAO (Reuse) & 9.88 & 51.73 & 11.76 & -- \\
         &  & SAO (CFG-matched) & \textbf{9.91} & \textbf{51.56} & \textbf{9.81} & -- \\
        \cmidrule(lr){2-7}
         & \multirow{3}{*}{4} & Random & $7.61{\scriptstyle\,\pm\,0.04}$ & $67.16{\scriptstyle\,\pm\,0.35}$ & $13.91{\scriptstyle\,\pm\,0.34}$ & -- \\
         &  & SAO (Reuse) & \textbf{9.40} & \textbf{54.38} & 15.01 & -- \\
         &  & SAO (CFG-matched) & 9.34 & 54.71 & \textbf{12.51} & -- \\
        \midrule
        \multirow{9}{*}{RLlamaGen} & \multirow{3}{*}{2} & Random & $7.17{\scriptstyle\,\pm\,0.07}$ & $72.39{\scriptstyle\,\pm\,0.74}$ & $11.84{\scriptstyle\,\pm\,0.10}$ & $30.89{\scriptstyle\,\pm\,0.01}$ \\
         &  & SAO (Reuse) & 8.21 & 63.27 & \textbf{11.61} & \textbf{30.93} \\
         &  & SAO (CFG-matched) & \textbf{8.23} & \textbf{63.13} & 11.88 & \textbf{30.93} \\
        \cmidrule(lr){2-7}
         & \multirow{3}{*}{3} & Random & $6.41{\scriptstyle\,\pm\,0.04}$ & $80.75{\scriptstyle\,\pm\,0.53}$ & $9.83{\scriptstyle\,\pm\,0.14}$ & $31.88{\scriptstyle\,\pm\,0.03}$ \\
         &  & SAO (Reuse) & 7.23 & 71.70 & \textbf{9.79} & 31.88 \\
         &  & SAO (CFG-matched) & \textbf{7.26} & \textbf{71.33} & 9.98 & \textbf{31.91} \\
        \cmidrule(lr){2-7}
         & \multirow{3}{*}{4} & Random & $5.82{\scriptstyle\,\pm\,0.05}$ & $88.83{\scriptstyle\,\pm\,0.82}$ & $9.90{\scriptstyle\,\pm\,0.07}$ & $\mathbf{32.30}{\scriptstyle\,\pm\,0.02}$ \\
         &  & SAO (Reuse) & 6.48 & 79.89 & \textbf{9.80} & \textbf{32.30} \\
         &  & SAO (CFG-matched) & \textbf{6.54} & \textbf{79.11} & 10.46 & \textbf{32.30} \\
        \bottomrule
    \end{tabular}%
\end{table*}

\section{Comparison with Alternative Generation Orders}
\label{sec:suppl_order_comparison}

We compare SAO with existing spatial and parallel-generation-inspired orders on ARPG-L and RLlamaGen to assess the benefit of constructing orders specifically for self-speculative decoding.
All methods use $K=16$, the same pretrained model, exact self-speculative decoder, and sampling configuration within each backbone; only the generation order changes.
Each complete order is determined before decoding and remains fixed throughout generation.
We consider three categories of alternative orders:

\begin{itemize}
    \item \textbf{Spatial scans and low-discrepancy orders.}
    We include Raster Scan, Spiral-in, Spiral-out, Z-curve, Subsample, and Alternate~\cite{esser2021taming}.
    Raster Scan visits positions row by row, while Spiral-in and Spiral-out follow spiral paths from the boundary to the center and in the reverse direction, respectively.
    Z-curve follows Morton order, Subsample recursively partitions the grid into interleaved spatial subgrids, and Alternate reverses the scan direction on successive rows.
    Halton~\cite{halton1960efficiency} maps a two-dimensional low-discrepancy sequence with bases 2 and 3 to the token grid, visiting each position at its first occurrence.

    \item \textbf{Prior-guided random orders.}
    Following the spatial priors studied in RandAR~\cite{pang2025randar}, we consider Hierarchical Random, Center-first Random, and Border-first Random.
    Hierarchical Random first visits the coarse subgrid containing every second row and column, followed by the remaining positions.
    Center-first Random prioritizes the central $8\times8$ region of the $16\times16$ grid, while Border-first Random prioritizes its complement.
    Each order independently shuffles positions within the prioritized region and within the remaining region, then concatenates the two permutations.

    \item \textbf{Parallel-generation-inspired orders.}
    PAR-inspired~\cite{wang2025par} partitions the grid into four equal regions and interleaves their corresponding local raster positions.
    LPD-inspired~\cite{zhang2026locality} combines proximity to previously selected positions with spatial separation within each group.
    It starts from a random position, prioritizes candidates with strong local proximity to the selected context, and uses spatially separated selections to complete each group.
    We adapt these ordering strategies to produce complete permutations for the same self-speculative decoder.
\end{itemize}

As shown in \cref{tab:order_comparison_arpg,tab:order_comparison_rllamagen}, decoding efficiency varies substantially across generation orders, even with the same pretrained model and self-speculative decoder.
Raster Scan, Spiral-in, Spiral-out, and Z-curve commit approximately four tokens per round on both backbones, whereas SAO commits 11.21 tokens on ARPG-L and 9.37 on RLlamaGen.
These differences highlight generation order as an important
dimension for optimizing self-speculative decoding.
Subsample and Halton also improve committed length and reduce NFE relative to Random, motivating a closer examination of their spatial patterns.

The visualizations in \cref{fig:generation_orders} reveal a spatial characteristic shared by Subsample, Halton, and SAO.
All three interleave earlier and later positions across the grid rather than filling contiguous regions sequentially.
Subsample exhibits a regular interleaved pattern, whereas Halton and SAO distribute generation ranks more irregularly.
This spatial organization may reduce dependencies among jointly drafted tokens, helping explain the strong performance of Subsample and Halton.
However, spatial dispersion alone does not capture the strength or direction of model-specific conditional dependencies, nor the cost of rejecting a token at different positions within a speculative window.
SAO incorporates both through the conditional-TV matrix and rejection-cost weighting, allowing it to distinguish orders with similar spatial dispersion but different predicted speculative efficiency.

SAO achieves the highest average committed length and the lowest NFE among all compared orders on both backbones.
Compared with Halton, the alternative with the lowest NFE, SAO increases $\mathbb{E}[L]$ from 10.46 to 11.21 on ARPG-L and from 8.79 to 9.37 on RLlamaGen.
The corresponding NFE decreases from 48.88 to 45.60 and from 59.26 to 55.54, respectively.
SAO also achieves the lowest FID on both backbones and the highest CLIP score on RLlamaGen.
These results demonstrate the benefit of constructing generation orders specifically for self-speculative decoding.

\section{Effect of Classifier-Free Guidance}
\label{sec:suppl_cfg}

We examine whether an SAO order constructed at CFG=1 remains effective at higher guidance scales.
At $K=16$, we evaluate CFG scales of 2, 3, and 4, comparing Random order with two SAO configurations.
\textbf{SAO (Reuse)} directly applies the order constructed at CFG=1, whereas \textbf{SAO (CFG-matched)} reconstructs the conditional-TV matrix and generation order at the corresponding guidance scale.
Within each guidance scale, all methods share the same decoder and sampling settings.

For guidance scale $g$, the conditional and unconditional logits are combined as~\cite{ho2022classifierfree}
\begin{equation}
\begin{aligned}
\ell_g(z\mid h)
={}&\ell_{\mathrm{uncond}}(z\mid h)\\
&+g\bigl[\ell_{\mathrm{cond}}(z\mid h)
-\ell_{\mathrm{uncond}}(z\mid h)\bigr].
\end{aligned}
\label{eq:suppl_cfg_logits}
\end{equation}
The two branches share the same generated visual-token history, while the conditioning input differs.
At temperature 1, the resulting sampling distribution is
\begin{equation}
p_{\theta,g}(z\mid h)
=\operatorname{softmax}\!\bigl(\ell_g(\cdot\mid h)\bigr)_z.
\label{eq:suppl_cfg_distribution}
\end{equation}
SAO (CFG-matched) uses this distribution to generate profiling histories, sample anchor values, and estimate conditional-TV influence:
\begin{equation}
\begin{aligned}
W^{(g)}_{u\rightarrow v}
={}&\mathbb{E}_{h\sim\mathcal{D}_g^{uv}}
\mathbb{E}_{a\sim p_{\theta,g}(x_u\mid h)}
\Big[\\
&\operatorname{TV}\bigl(
 p_{\theta,g}(x_v\mid h),\\
&\hspace{3em}p_{\theta,g}(x_v\mid h\oplus(u,a))
\bigr)\Big].
\end{aligned}
\label{eq:suppl_cfg_matrix}
\end{equation}
Here, $\mathcal{D}_g^{uv}$ is the distribution of profiling histories generated at guidance scale $g$ in which neither $u$ nor $v$ has been generated.
We retain the same random-order history sampling protocol, 5K-trajectory budget, prefix-length sampling, and anchor-position sampling used at CFG=1.
The rejection-cost objective and conditional-expectation construction remain unchanged, with $W^{(g)}$ replacing the original matrix.
Thus, SAO (Reuse) decodes at guidance scale $g$ using the order constructed from $W^{(1)}$, while SAO (CFG-matched) uses the order constructed from $W^{(g)}$.

As shown in \cref{tab:cfg_comparison}, SAO (Reuse) achieves decoding efficiency close to SAO (CFG-matched) across all evaluated guidance scales on both backbones.
Both increase average committed length and reduce NFE relative to Random, while CFG-specific profiling provides little additional efficiency benefit.

These results suggest that the ordering preferences captured at CFG=1 remain useful despite changes in the sampling distributions at higher guidance scales.
Consequently, reusing the order constructed at CFG=1 avoids additional profiling while achieving decoding efficiency close to CFG-matched construction over the tested guidance scales.

\end{document}